\documentclass{article}

\usepackage[a4paper]{geometry}
\usepackage{amsfonts,amsmath,amssymb} 
\usepackage{mathtools}
\usepackage{url}

\usepackage{booktabs}
\usepackage{multirow}

\usepackage{rotating}

\usepackage{bm}
\usepackage{eurosym}
\usepackage{graphicx}
\usepackage{hyperref}
\hypersetup{colorlinks,allcolors=black}
\usepackage{lineno}
\usepackage{listings}
\usepackage{multirow}
\usepackage{url}
\usepackage{xcolor}
\usepackage{xspace}
\usepackage[colorinlistoftodos]{todonotes}

\newcommand{\aggr}{@}
\newcommand{\mycomment}[1]{\vspace*{1em} \parbox{0.9\textwidth}{\color{blue}{COMMENT: \newline #1 \newline }}}

\usepackage[english]{babel}
\usepackage{lscape}
\usepackage{textcomp}

\usepackage{multirow}
\usepackage{stmaryrd}
\usepackage[ruled]{algorithm}
\usepackage[noend]{algpseudocode}
\usepackage{pbox}
\usepackage{listings}
\floatname{algorithm}{Algorithm}

\usepackage{verbatim}

\newcommand{\ceil}[1]{\lceil #1 \rceil}

\usepackage{tikz}
\usetikzlibrary{trees}
\usetikzlibrary{shadows}
\usetikzlibrary{arrows}

\usepackage{rotating,makecell}
\usepackage{soul}
\definecolor{maize}{rgb}{0.98, 0.93, 0.37}
\definecolor{green-yellow}{rgb}{0.68, 1.0, 0.18}
\definecolor{operamauve}{rgb}{0.72, 0.52, 0.65}
\definecolor{paleblue}{rgb}{0.69, 0.93, 0.93}

\makeatletter
\usepackage{etoolbox}
  \patchcmd{\@savemarbox}{\color@vbox}{\color@vbox\normalcolor}{}{}
\makeatother
\newcommand{\fuzzyowlpnrule}{\textsc{PN-OWL}}
\newcommand{\fuzzyowlpnruleplus}{$\textsc{PN-OWL}^+$}

\floatname{algorithm}{Algorithm}

\newcommand{\nd}{\noindent}
\newcommand{\ind}{\mathsf{I}}
\newcommand{\indkb}{\ind_{\KB}}

\newcommand{\inds}[1]{\ind_{#1}}

\newcommand{\unlabel}{\mathcal{E}^u}
\newcommand{\posi}{\mathcal{E}^+}
\newcommand{\nega}{\mathcal{E}^-}
\newcommand{\train}{\mathcal{E}_{train}}
\newcommand{\test}{\mathcal{E}_{test}}
\newcommand{\refalgo}[1]{Algorithm~\ref{#1}}

\newtheorem{remark}{Remark}

\newtheorem{example}{Example}[section]

\graphicspath{{./}{images/}}
\newcommand{\m}{\alpha}

\newcommand{\unit}{[0,1]}
\newcommand{\unitp}{(0,1]}

\newcommand{\dt}{{\mathbf{d}}}

\newcommand{\bed}[2]{bed(#1, #2)}

\newcommand{\ins}{\,{\in}\,}

\newcommand{\DL}{\ensuremath{\mathcal{DL}}}
\newcommand{\T}{\ensuremath{\mathcal{T}}}
\newcommand{\A}{\ensuremath{\mathcal{A}}}
\newcommand{\K}{\ensuremath{\mathcal{K}}}
\newcommand{\I}{\ensuremath{\mathcal{I}}\xspace}      

\newcommand{\EL}{\ensuremath{\mathcal{EL}}}

\newcommand{\ALCAGGR}{\ensuremath{\mathcal{ALC}_\aggr}}

\newcommand{\dllite}{\mbox{DL-Lite}}

\newcommand{\D}{{\mathbf{D}}}
\newcommand{\eqs}{\,{=}\,}

\newcommand{\calS}{{\cal S}}
\newcommand{\E}{{\cal E}}

\newcommand{\alc}{\mathcal{ALC}}

\newcommand{\alcaggr}{\ALCAGGR}
\newcommand{\el}{\mathcal{EL}}

\newcommand{\eld}{\el(\D)}

\newcommand{\andc}{\sqcap}
\newcommand{\all}{\forall}
\newcommand{\some}{\exists}

\newcommand{\notc}{\neg}
\newcommand{\orc}{\sqcup}
\newcommand{\csome}{\exists}

\newcommand{\bottomc}{\perp}
\newcommand{\topc}{\top}
\newcommand{\impc}{\sqsubseteq}
\newcommand{\highi}[1]{{#1}^{\cal I} }

\newcommand{\calC}{\ensuremath{\mathcal{C}} }
\newcommand{\calE}{\ensuremath{\mathcal{E}} }

\newcommand{\tuple}[1]{\langle #1 \rangle }

\newcommand{\notf}{\ominus}%{\neg}
\newcommand{\orf}{\oplus}%{\lor}
\newcommand{\andf}{\otimes}%{\land}
\newcommand{\impf}{\Rightarrow}

\newcommand{\fuzzyg}[2]{\mbox{$\tuple{#1,#2}$}}
\newcommand{\rimp}{\rightarrow}

\newcommand{\foil}{\textsc{Foil}}
\newcommand{\pfoil}{p\textsc{Foil}}

\newcommand{\foildl}{\foil-\DL}

\newcommand{\foildlpn}{pn\foildl}
\newcommand{\pfoildl}{\pfoil-\DL}
\newcommand{\fuzzyowladaboost}{\textsc{Fuzzy OWL-Boost}}

\algblockdefx[function]{FUNCTION}{ENDFUNCTION}[2]{\textbf{function} \texttt{#1(#2)}}{\textbf{end}}
\algblockdefx[returnCondition]{IfReturn}{EndIfReturn}[1]{\textbf{if}(#1)}
[1]{\indent \textbf{return} #1;}

\newcommand{\ie}{{\em i.e. }}
\newcommand{\eg}{{\em e.g. }}
\newcommand{\cf}{{\em cf. }}
\newcommand{\viz}{{\em viz.}}
\newcommand{\wrt}{{w.r.t.}}

\newcommand{\cass}[2]{\mbox{$#1$:$#2$}}
\newcommand{\rass}[3]{\mbox{$(#1,#2)$:$#3$}}

\newcommand{\ii}[1]{\emph{(#1)}}

\newcommand{\KB}{{\K}}

\title{\fuzzyowlpnrule: A Two Stage Algorithm to Learn Fuzzy Concept Inclusions from OWL Ontologies}

\author{Franco Alberto Cardillo \\
CNR-ILC \\
Pisa, Italy \\
\and Franca Debole\\
CNR-ISTI \\
Pisa, Italy \\
\and Umberto Straccia\\
CNR-ISTI \\
Pisa, Italy 
}

\begin{document}
\maketitle

%\begin{frontmatter}
%
%%\title{\fuzzyowlpnrule: Learning Fuzzy Concept Inclusion Axioms from OWL Ontologies  via $\mathbb{R}$eal AdaBoost}
%\title{\fuzzyowlpnrule: Learning Fuzzy Concept Inclusions via Real-Valued Boosting}
%% Order of authors to be revised
%\author[label1]{Franco Alberto Cardillo}
%\ead{francoalberto.cardillo@ilc.cnr.it}
%
%\author[label2]{Umberto Straccia}
%\ead{umberto.straccia@isti.cnr.it}
%
%\address[label1]{ILC - CNR, Pisa}
%\address[label2]{ISTI-CNR, Pisa, Italy}

%--------------------------------------------------------------------------------------------------------------------------------------------------------------------

\begin{abstract}
OWL ontologies are a quite popular way to describe structured knowledge in terms of classes, relations among classes and class instances.  

In this paper, given a  target class $T$ of an OWL ontology, with a focus on ontologies with real- and boolean-valued data properties, we address the problem of learning graded fuzzy concept inclusion  axioms with the aim of describing enough conditions for being an individual classified as instance of the class $T$. 

To do so, we present \fuzzyowlpnrule~that is a two-stage learning algorithm made of a P-stage and an N-stage. Roughly, in the P-stage the algorithm tries to cover as many positive examples as possible (increase \emph{recall}), without compromising too much \emph{precision}, while in the N-stage, the algorithm tries to rule out as many false positives, covered by the P-stage, as possible.
%
% Informally, the basic principle of PN-rule consists of a \emph{P-stage} in which \emph{positive} rules are learnt that cover most positive instances of a target class, while keeping the non-positive rate at a reasonable level, and an \emph{N-stage} in which \emph{negative} rules are learnt to remove most of the non-positive covered by the P-stage. The two rule sets are then used to build up a decision method to classify an object being instance of the target class or not.
%
\fuzzyowlpnrule~then aggregates the fuzzy inclusion axioms learnt at the P-stage and the N-stage by combining them via aggregation functions to allow for a final decision whether an individual is instance of $T$ or not.

We also illustrate its effectiveness by means of an experimentation. An interesting feature is that fuzzy datatypes are built automatically, the learnt fuzzy concept inclusions can be represented directly into Fuzzy OWL 2 and, thus, any Fuzzy OWL 2 reasoner can then be used to automatically determine/classify (and to which degree) whether an individual belongs to the target class $T$ or not.
\end{abstract}

%\begin{keyword}
%OWL 2 Ontologies \sep Machine Learning \sep Real-valued AdaBoost \sep Fuzzy Logic \sep Concept Inclusion Axioms
%\end{keyword}
%
%\end{frontmatter}

%%%%%%%%%%%%%%%%%%%%%%%%%%%%%%%%%%%%%%%%%%%%%%%%%%%%%%%%%%%%%%%%%%%%%%%%%%%%%%%%%%%%%%%%%%%%%%%%%

\section{Introduction}
\nd OWL 2 ontologies~\cite{OWL2} are a popular means to represent \emph{structured} knowledge and its formal semantics is based on \emph{Description Logics} (DLs)~\cite{Baader07}. The basic ingredients of DLs are concept descriptions, inheritance relationships among them and instances of them.

In this work, we  focus on the problem of automatically learning fuzzy $\EL(\D)$ concept inclusion axioms from OWL 2 ontologies based on the terminology and instances within it. Despite an important amount of work has been carried about DLs, the application of machine learning techniques to OWL 2 ontologies is relatively less addressed compared to the \emph{Inductive Logic Programming} (ILP) setting (see \eg~\cite{deRaedt08,DeRaedt17} for more insights on ILP). We refer the reader to~\cite{Lisi19,Rettinger12} for an  overview and to Section~\ref{sec:relatedWork}. 

In this paper, the problem we address is the following: given a target class $T$ of an OWL ontology,  learn rule-like graded fuzzy $\EL(\D)$~\cite{Bobillo11c,Bobillo18,Straccia13} concept inclusion axioms with the aim of describing sufficient conditions for being an individual classified as instance of the class $T$.

The following example illustrates the problem we are going to address.\footnote{See also \eg~\cite{Cardillo22,Lisi13a,Lisi15,Straccia15} for an analogous example.}

\begin{example} \label{runes}
Consider an ontology~\cite{Buehmann16,Cardillo22} that describes the meaningful entities of mammography image analysis. An excerpt of this ontology is given in Fig.~\ref{mammonto}.
\begin{figure}
\begin{center}
\includegraphics[scale=0.15]{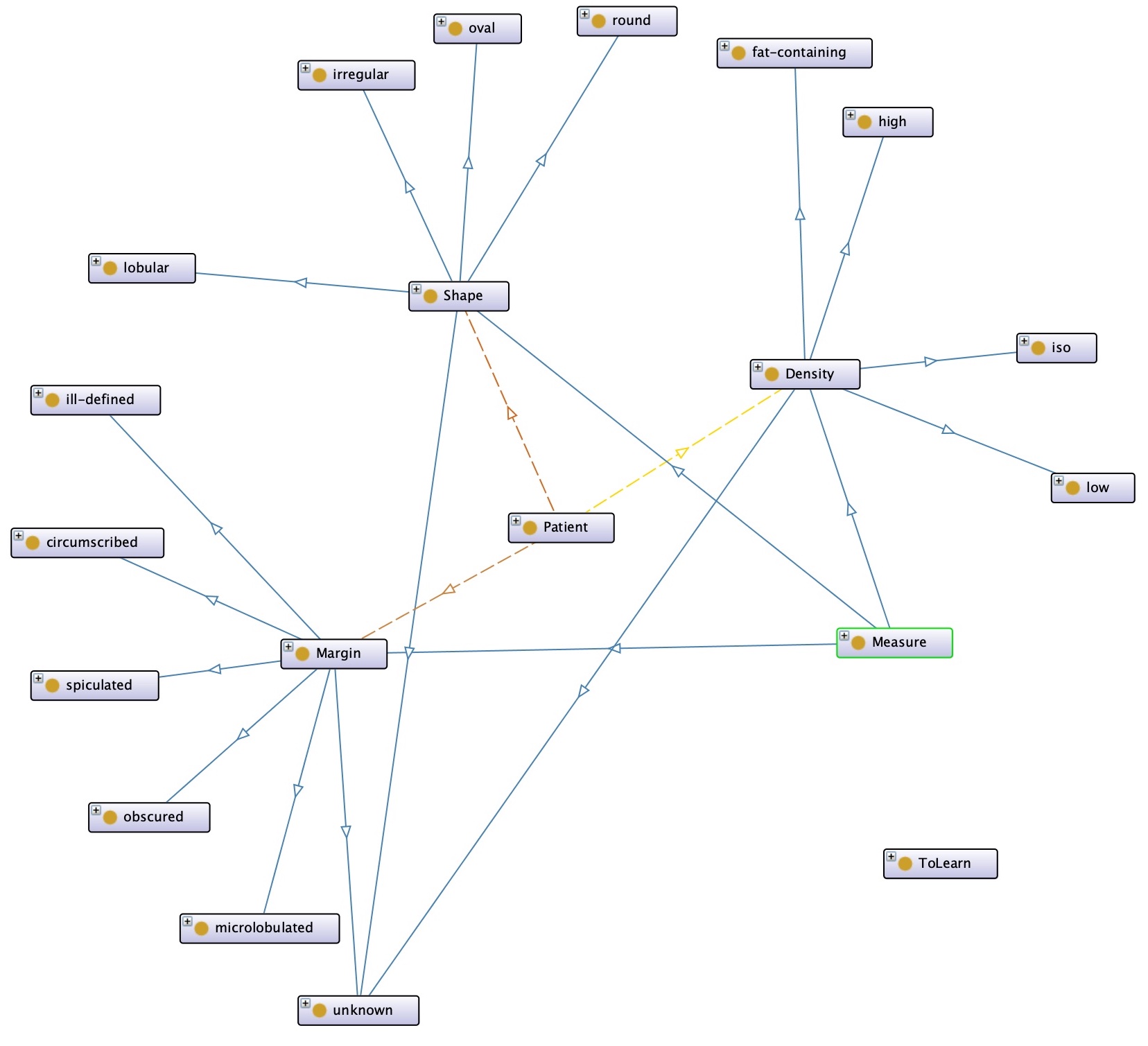}
\end{center}
  \caption{Excerpt of the mammographic ontology.}
  \label{mammonto}
\end{figure}
Now, suppose we have a set of patients that exhibit a cancer (\emph{positive} examples) and another set which does not (\emph{negative} examples). Now, one may ask about what characterises the patients with cancer (our target class $T$). Then one may learn from the ontology the following fuzzy $\EL(\D)$ concept inclusion 
(expressed in the so-called Fuzzy OWL 2 syntax~\cite{Bobillo11c})\footnote{\url{http://www.umbertostraccia.it/cs/software/fuzzyDL/fuzzyDL.html}}

\begin{quote}
{\tt
({\bf implies} ({\bf and} ({\bf some} hasDensity  fat-containing)  ({\bf some} hasMargin  spiculated) ({\bf some} hasShape irregular)  ({\bf some} hasAge hasAge\_high))  Cancer $0.86$) 
} ,
\end{quote}

\nd where the fuzzy set {\tt hasAge\_high} is defined as 
\begin{quote}
{\tt 
({\bf define-fuzzy-concept} hasAge\_high right- (0,150,60,80))
}
\end{quote}

% (expressed in the so-called Manchester syntax~\cite{OWL2Manchester})

% \begin{quote}
% $\langle$ (hasDensity {\bf some} fat-containing) {\bf and} (hasMargin {\bf some} spiculated) {\bf and} (hasShape {\bf some} irregular) {\bf and} (hasAge {\bf some} hasAge\_high) SubClassOf Cancer, $0.86$ $\rangle$.
% \end{quote}

\nd In words, 

\begin{quote}
``if the density is fat-containing, the margin is spiculated, the shape is irregular and the age is high then it is cancer, with confidence $0.86$".
\end{quote}
%(see, \eg~\cite{Cardillo22} and also Section~\ref{sec:eval} later on). %\qed
\end{example}

\nd In this work, the objective is the same as in \eg~\cite{Cardillo22,Lisi15,Straccia15} except that now we propose to rely on an adaptation of the PN-rule~\cite{Agarwal00,Agarwal01,Joshi01,Joshi02} algorithm to the (fuzzy) OWL case. Further, like in~\cite{Lisi13a,Straccia15}, we continue to support so-called \emph{fuzzy concept descriptions} and \emph{fuzzy concrete domains}~\cite{Lukasiewicz08a,Straccia05d,Straccia13}, such as the expression {\tt ({\bf some} hasAge  hasAge\_high)} (\viz~an aged person) in Example~\ref{runes} above, which is a fuzzy concept, \ie~a concept for which the belonging of an individual to the class is not necessarily a binary yes/no question, but rather a matter of (truth) degree in $[0,1]$. 

For instance, in our example, the degree depends on the person's age: the higher the age the older is the person, \eg~modelled via a so-called \emph{right shoulder function} (see Figure~\ref{fig:muf}(d)).
Here, the range of the `attribute' {\tt hasAge} becomes a so-called fuzzy concrete domain~\cite{Straccia05d,Straccia13}.

Let us recap that the basic principle of PN-rule consists of a \emph{P-stage} in which \emph{positive} rules (called \emph{P-rules}) are learnt to cover as many as possible instances of a target class, and keeping the non-positive rate at a reasonable level, and an \emph{N-stage} in which \emph{negative} rules (called \emph{N-rules}) are learnt to remove most of the non-positive examples covered by the P-stage. The two rule sets are then used to build up a decision method to classify an object being instance of the target class or not~\cite{Agarwal00,Agarwal01,Joshi01,Joshi02}. It is worth noting that what differentiates this method from all others is its second stage. It learns N-rules that essentially remove the non-positive examples (so-called false positives) collectively covered by the union of all the P-rules. 

The following are the main features of our two stage algorithm, called \fuzzyowlpnrule:
\begin{itemize}
\item at the P-stage, it generates a set of fuzzy $\el(\D)$ inclusion axioms, the P-rules, that cover as many as possible instances of a target class $T$ without compromising too much the amount on non-positives (\ie, try to increase the so-called \emph{recall}); 

\item at the N-stage, it generates a set of fuzzy $\el(\D)$ inclusion axioms, the N-rules, that cover as many as possible of non-positive instances of class $T$ (\ie, then try to increase the so-called \emph{precision});
%covered by the P-stage;

\item the fuzzy inclusion axioms are then combined (aggregated) into a new fuzzy inclusion axiom describing sufficient conditions for being an individual classified as an instance of the target class $T$ (\ie~the combination aims at increasing the overall effectiveness, \eg~the so-called F1-measure);

\item all fuzzy inclusion axioms may possibly include fuzzy concepts and fuzzy concrete domains, where each axiom has a leveraging weight (specifically, called \emph{confidence} or \emph{precision});

\item all generated fuzzy concept inclusion axioms can  be directly encoded as \emph{Fuzzy OWL 2} axioms~\cite{Bobillo10,Bobillo11c}. Therefore, a Fuzzy OWL 2 reasoner, such as \emph{fuzzyDL}~\cite{Bobillo08a,Bobillo16}, can then be used to automatically determine (and to which degree) whether an individual belongs to the target class $T$.
\end{itemize}

\nd We will  illustrate the effectiveness of \fuzzyowlpnrule~by means of an experimentation that shows that the effectiveness of the combined approach increases \wrt~a baseline based on the P-stage only.

In the following, we proceed as follows: in Section~\ref{sec:background}, for the sake of completeness, we recap the salient notions we will rely on this paper. Then, in Section~\ref{sec:learn} we will present our algorithm \fuzzyowlpnrule, which is evaluated in Section~\ref{sec:eval}. In Section~\ref{sec:relatedWork} we compare our work with closely related work appeared so far. 
%For completeness, we refer to~\ref{sec:appendixLearning} in which we provide a much more extensive list of references related to  OWL rule learning, though less related to our setting. 
Section~\ref{sec:conclusions} concludes and points to some topics of further research. 

%%%%%%%%%%%%%%%%%%%%%%%%%%%%%%%%%%%%%%%%%%%%%%%%%%%%%%%%%%%%%%%%%%%%

%%%%%%%%%%%%%%%%%%%%%%%%%%%%%%%%%%%%%%%%%%%%%%%%%%%%%%%%%%%%%%%%%%%%

\section{Background} \label{sec:background}

\nd We introduce the main notions related to \emph{(Mathematical) Fuzzy Logics} and  \emph{Fuzzy Description Logics} we will use in this work (see also~\cite{Bobillo18,Straccia13}).

%---------------------------------
\paragraph{Mathematical Fuzzy Logic.} \label{sect:fuzzy-logic}
%\subsection{Mathematical Fuzzy Logic} \label{sect:fuzzy-logic}
%\vspace{1ex}
%\nd {\bf Mathematical Fuzzy Logic.}
%---------------------------------
%
%\vspace{-4ex}
\begin{figure}
%\begin{scriptsize}
\begin{center}
\begin{tabular}{c@{\ \ \ \ \ \ \ \ }c@{\ \ \ \ \ \ \ \ }c@{\ \ \ \ \ \ \ \ }c}
\includegraphics[scale=0.3]{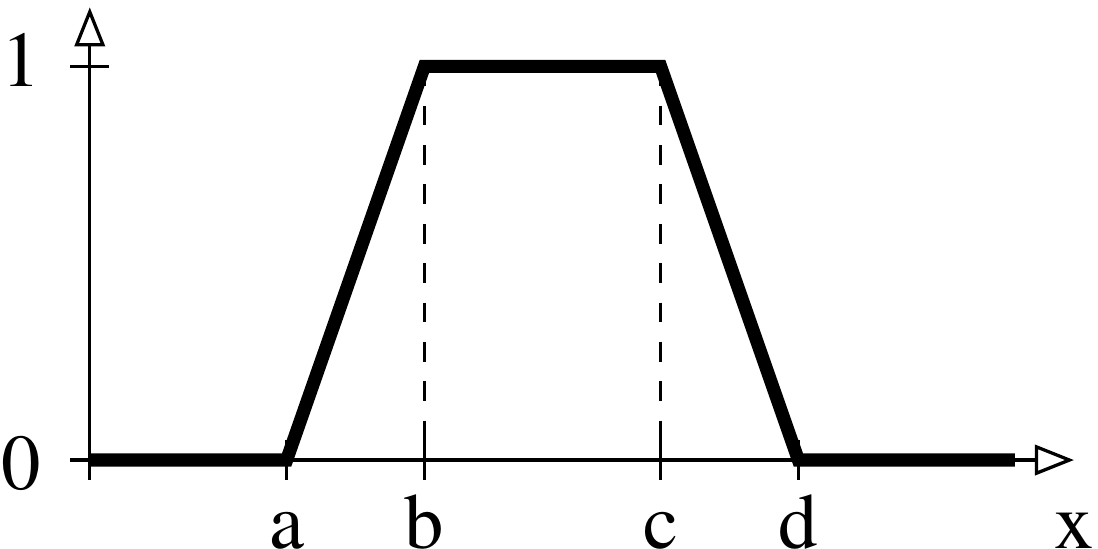} &
\includegraphics[scale=0.3]{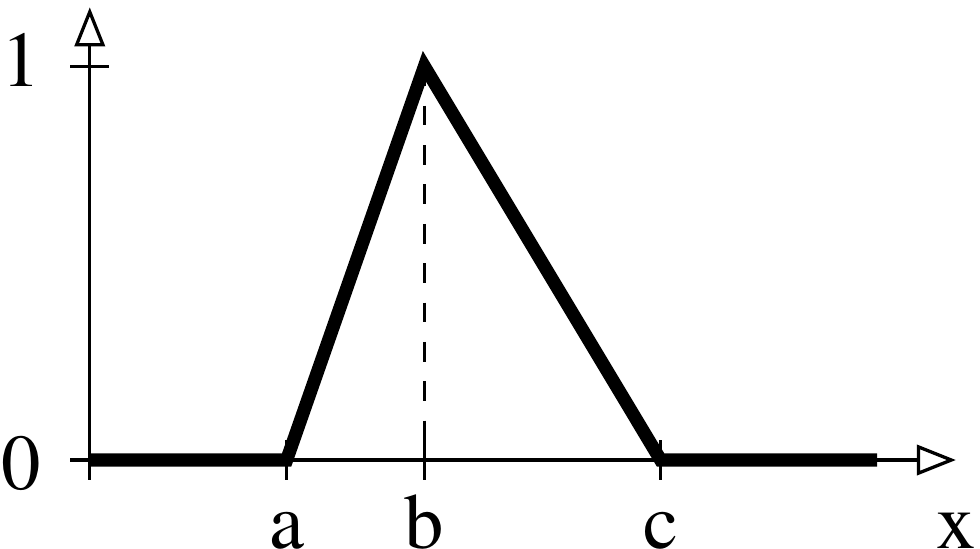} & \\
(a) & (b)   \\
\includegraphics[scale=0.3]{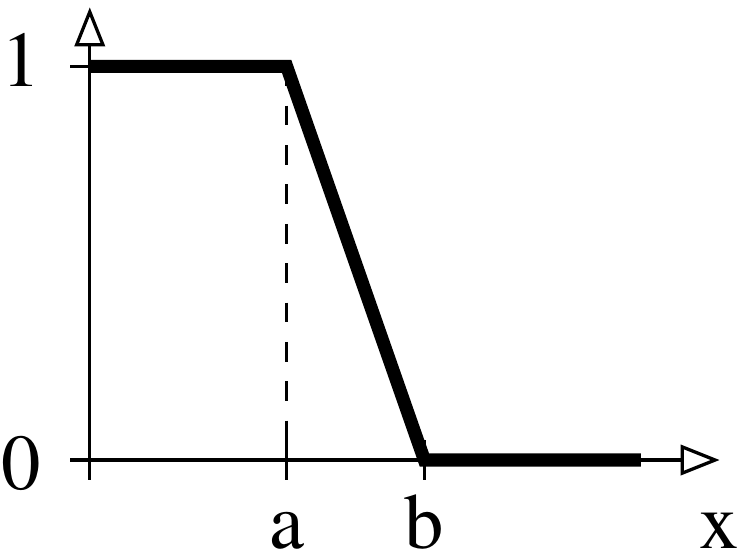} &
\includegraphics[scale=0.3]{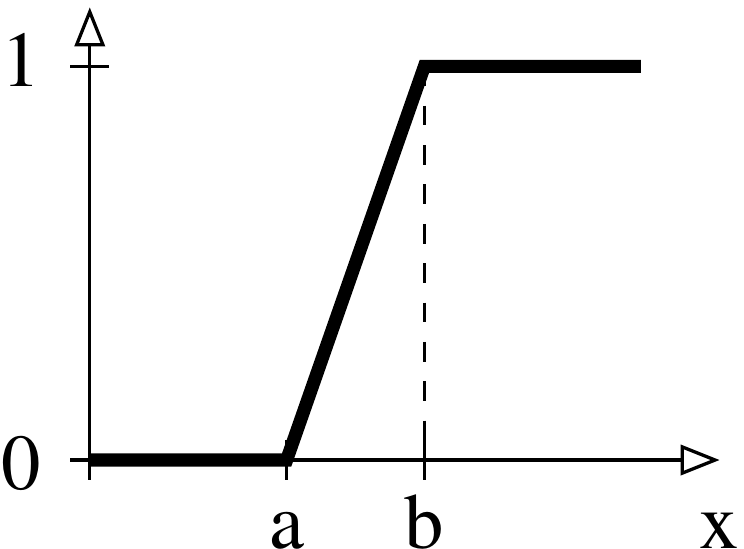} \\
(c) & (d)
%\\
%\includegraphics[scale=0.175]{lfunction.pdf} &
%\includegraphics[scale=0.175]{gammafunction.pdf} \\
%(c) & (d)
\end{tabular}
%\vspace*{-1ex}
\caption{(a) Trapezoidal function $\mathit{trz}(a,b,c,d)$,
(b) triangular function $\mathit{tri}(a,b,c)$, (c) left shoulder
function~$\mathit{ls}(a,b)$, and (d) right shoulder function $\mathit{rs}(a,b)$.}
\label{fig:muf}
%\vspace*{-5ex}
%\vspace{-.5cm}
\end{center}
%\end{scriptsize}
\end{figure}
%\vspace{-5ex}
%
%
%
\nd \emph{Fuzzy Logic} is the logic of \emph{fuzzy sets}~\cite{Zadeh65}. A \emph{fuzzy set} $A$ over a countable crisp set $X$ is a function $A \colon X \to [0,1]$, called \emph{fuzzy membership} function of $A$. A \emph{crisp set} $A$ is characterised by  a membership function   $A \colon X \to \{0,1\}$ instead. Often, fuzzy set operations conform to 
$(A \cap B)(x) = \min(A(x), B(x))$, $(A \cup B)(x) = \max(A(x), B(x))$ and $\bar{A}(x) = 1- A(x)$ ($\bar{A}$ is the set complement of $A$), 
the \emph{cardinality} of a fuzzy set is defined as  $|A| = \sum_{x\in X} A(x)$, while 
the \emph{inclusion degree} between $A$ and $B$ is defined as $deg(A,B) = \frac{|A \cap B|}{|A|}$. 

The trapezoidal, the triangular, the left-shoulder function, and the right-shoulder
function are frequently used to specify membership functions of fuzzy sets (see Figure~\ref{fig:muf}).

% Although fuzzy sets have a  greater expressive power than
% classical crisp sets, their usefulness depends critically on the
% capability to construct appropriate membership functions for various
% given concepts in different contexts.  We refer the
% interested reader to, \eg,~\cite{Klir95}. 

\begin{figure}
\begin{center}
\begin{tabular}{cc}
\includegraphics[scale=0.33]{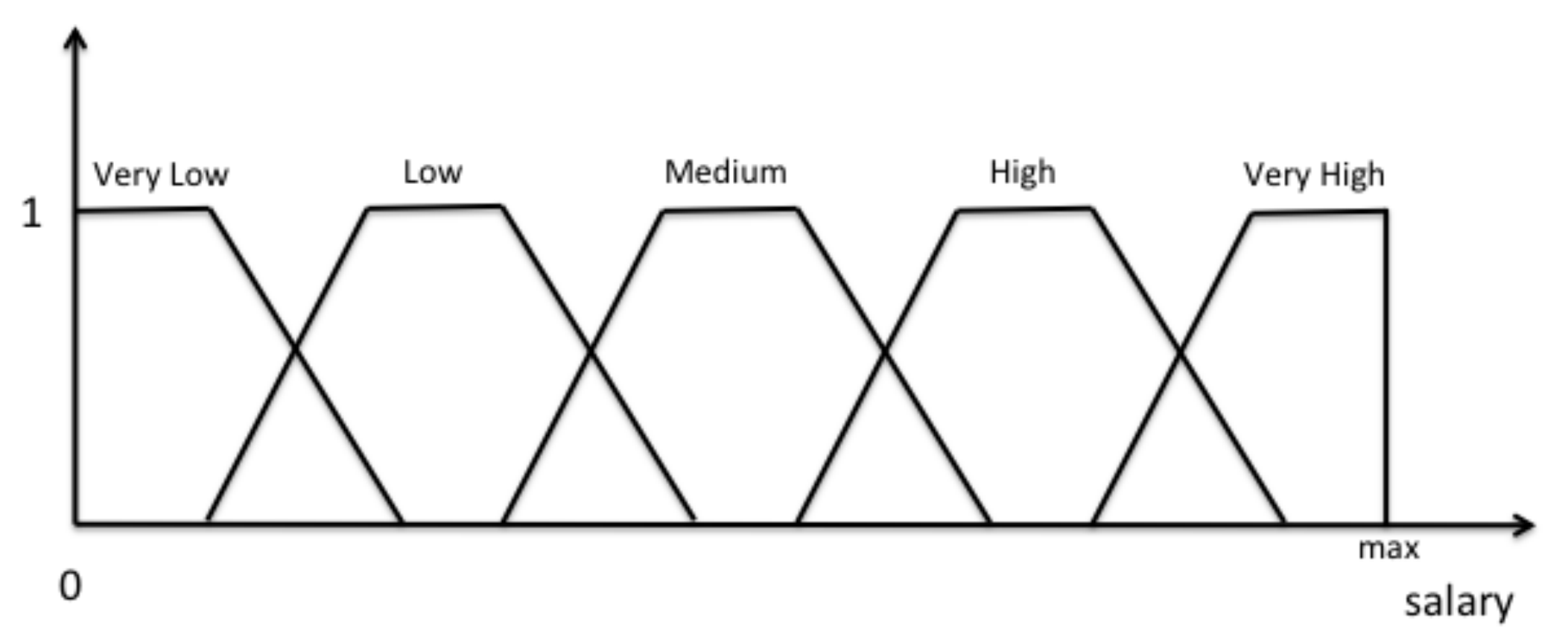} &
\includegraphics[scale=0.3]{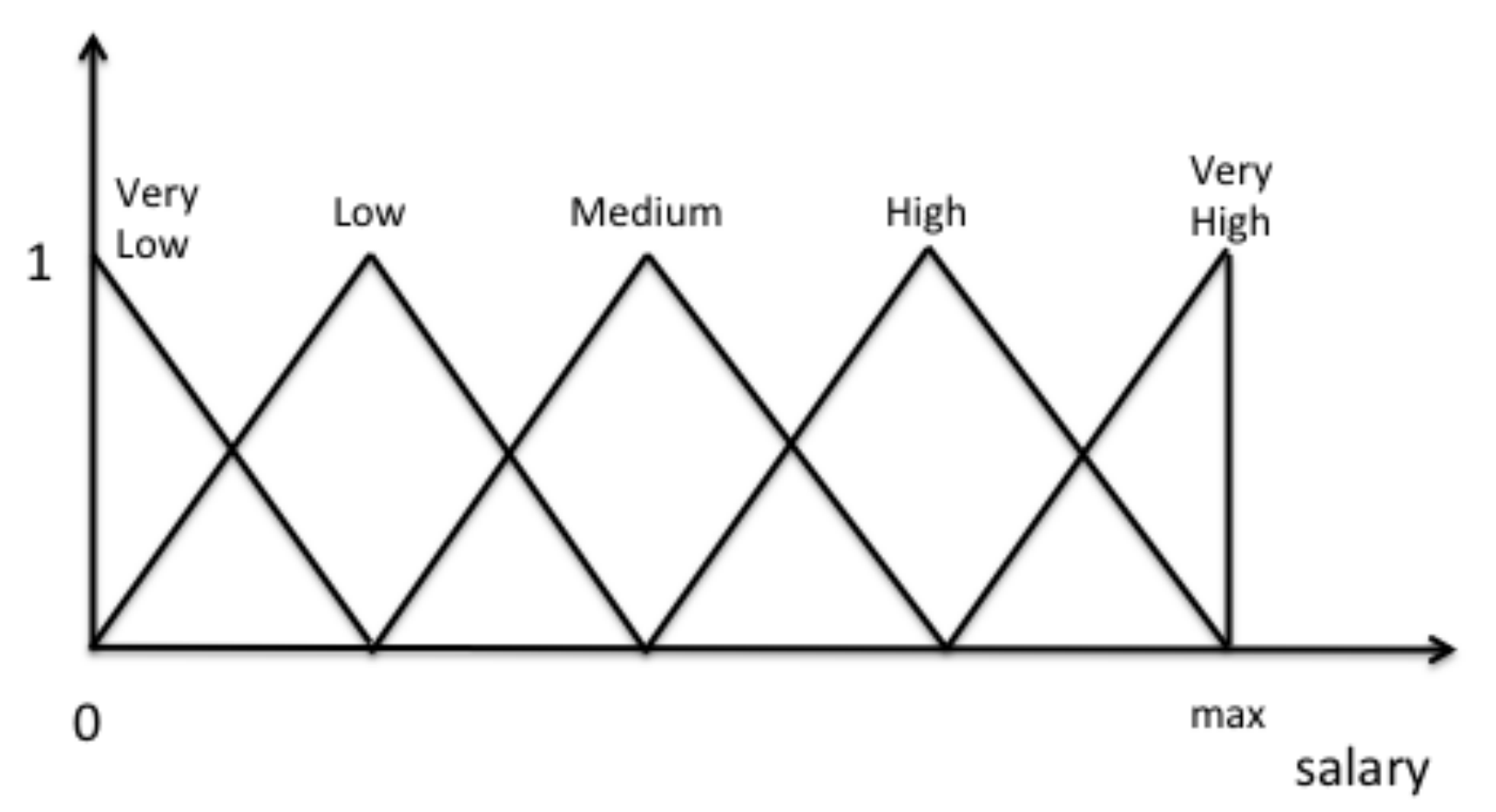}
\end{tabular}
%\vspace*{-4ex}
\caption{Uniform fuzzy sets over salaries: trapezoidal (left) or triangular (right).}
\label{partfuzzytrz}
%\vspace*{-2ex}
\end{center}
\end{figure}

One easy and typically satisfactory method to define the membership
functions is to uniformly partition the range of, \eg~person's age values
(bounded by a minimum and maximum value), into 3, 5 or 7 fuzzy sets
using triangular (or trapezoidal) functions (see Figure~\ref{partfuzzytrz}). Another popular approach may consist in
using the so-called \emph{c-means} fuzzy clustering algorithm (see, \eg~\cite{Bezdek81}) with 3,5 or 7 clusters, 
where the fuzzy membership functions are triangular functions built around the centroids of the clusters (see also~\cite{Huitzil18}).

%
% The latter
%is the more used one, as it has less parameters and is also the approach we adopt.
%Of course, there are infinitely many alternatives to the standard fuzzy set operations illustrated here (see \eg~\cite{Klir95}).
%\vspace{-4ex}
%\begin{figure}
%\begin{center}
%\includegraphics[scale=0.4]{partfuzzytri}
%\caption{Fuzzy sets over salaries using triangular functions.}\label{partfuzzytri}
%\end{center}
%\end{figure}
%\vspace{-6ex}

In \emph{Mathematical Fuzzy Logic}~\cite{HajekP98}, the convention
prescribing that a formula $\phi$ is either true or false (\wrt~an interpretation $\I$) is changed and
is a matter of degree measured on an ordered scale that is no longer
$\{0, 1\}$, but typically $[0,1]$. This degree is called {\em degree of
truth}  of the formula $\phi$ in the interpretation $\I$.
A \emph{fuzzy formula} has the form
$\fuzzyg{\phi}{\m}$, where $\m \ins (0,1]$ and $\phi$ is a
First-Order Logic (FOL) formula, encoding that the degree of truth  of $\phi$ is {\em
greater than or equal to $\m$}. 
% So, for instance, $\fuzzyg{Cheap(HotelVerdi)}{0.8}$ states that `Hotel Verdi is cheap' is true to degree greater or equal $0.8$.
%
%Usually, the \emph{truth space} is $L = [0,1]$. Another popular
%truth space is the finite truth space $\unitn  = \{0, \frac{1}{n-1},
%\ldots, \frac{n-2}{n-1}, 1\}$ for some natural number $n> 1$. Of
%course, $L_{2}$ is the usual classical two-valued case.
%%With $\unitp$ we denote $\unit \setminus \{0 \}$.
%
From a semantics point of view, a \emph{fuzzy interpretation} $\I$ maps each atomic formula 
into $[0,1]$ and is then extended inductively to all FOL formulae as follows:
%{\small
\begin{eqnarray*}
\I(\phi \land \psi)  & = &  \I(\phi) \otimes \I(\psi) \ \ , \ \ \I(\phi \lor \psi) \ = \  \I(\phi)\oplus \I(\psi)\\
 \I(\phi \rightarrow \psi) & = & \I(\phi) \impf \I(\psi) \ \ , \ \  \I(\neg\phi) \  = \ {\ominus}\,\I(\phi) \\
\I(\exists x. \phi(x)) &  = &  \sup_{y \in \highi{\Delta}}\I(\phi(y)) \ \ , \ \  \I(\forall x. \phi(x))  \ = \ \inf_{y \in \highi{\Delta}}\I(\phi(y)) \ ,
\end{eqnarray*}
%}
\nd where $\highi{\Delta}$ is the (non-empty) domain of $\I$, and $\otimes$,
$\oplus$, $\impf$, and $\ominus$ are so-called \emph{t-norms},
\emph{t-conorms}, \emph{implication functions}, and \emph{negation
functions}, respectively, which extend the Boolean conjunction,
disjunction, implication, and negation, respectively, to the fuzzy case.

%Table~\ref{tab:ImplicationFunctions} (see~\cite{Klir95} for more and properties of them) recalls the names of popular implication functions.
%\begin{table}[t]
%\begin{center}
%%{\small
%\begin{tabular}{|l|c|} \hline
%\textbf{Name} \ \ \ & \textbf{$x \impf y$} \\ \hline
%G{\"o}del & $\left\{
%\begin{array}{ll}
%1 & \mathrm{if \ } x \leq y \\
%y & \mathrm{if \ } x > y
%\end{array} \right .$ \\ \hline
%Gougen & $\left\{
%\begin{array}{ll}
%1 & \mathrm{if \ } x \leq y \\
%y/x & \mathrm{if \ } x > y
%\end{array} \right .$ \\ \hline
%Kleene-Dienes & $\max(1-x,y)$ \\ \hline
%Lukasiewicz & $\min(1, 1-x+y)$ \\ \hline
%Reichenbach & $1-x + xy$ \\ \hline
%\end{tabular}
%%}
%\end{center}
%\caption{Implication functions.}
%\label{tab:ImplicationFunctions}
%\end{table}
%\vspace*{1ex}

One usually distinguishes three different logics, namely \L{}ukasiewicz, G{\"o}del, and Product
logics~\cite{HajekP98},\footnote{Notably, a theorem states that any other continuous t-norm can be obtained as a combination of them.} whose truth combination functions are reported in Table~\ref{tab:norm}.

%\vspace{-2ex}
\begin{table}
\caption{Truth combination functions for fuzzy logics.}
\label{tab:norm}
%\vspace*{-3ex}
{\footnotesize
%{\scriptsize
%{\tiny
%{\small
\begin{center}{
\begin{tabular}{|c||ccc|} \hline
     & \mbox{\L{}ukasiewicz} &  \mbox{G{\"{o}}del}  & \mbox{Product}    \\ \hline
    $\m_1  \otimes \m_2$ & $\max(\m_1+\m_2-1,0)$ & $\min(\m_1,\m_2)$ & $\m_1\cdot \m_2$  \\ [2.5ex]
    $\m_1 \oplus \m_2$ &  $\min(\m_1+\m_2,1)$ & $\max(\m_1,\m_2)$ & $\m_1+\m_2-\m_1\cdot \m_2$   \\ [2.5ex]
    $\m_1 \impf \m_2$ & $\min(1-\m_1+\m_2,1)$ &
    $\begin{cases} 1 & \mbox{\!if } \m_1\leq \m_2  \\
     \m_2 & \mbox{\!otherwise}\end{cases}$
     & $\begin{cases} 1 & \mbox{\!if } \m_1\leq \m_2  \\
     \m_2/\m_1 & \mbox{\!otherwise}\end{cases}$ \\ [5ex]
    ${\ominus}\, \m$ & $1-\m$ &
    $\begin{cases} 1 & \mbox{\!if } \m=0\\
                                 0 & \mbox{\!otherwise}\end{cases}$
     & $\begin{cases} 1 & \mbox{\!if } \m=0\\
                                 0 & \mbox{\!otherwise}\end{cases}$ \\ [5ex] \hline
\end{tabular}}
%
% \begin{tabular}{c||ccc|c} \hline
%      & \mbox{\L{}ukasiewicz} &  \mbox{G{\"{o}}del}  & \mbox{Product}  & \mbox{`typical'}  \\ \hline
%     $\m_1  \otimes \m_2$ & $\max(\m_1+\m_2-1,0)$ & $\min(\m_1,\m_2)$ & $\m_1\cdot \m_2$ & $\min(\m_1,\m_2)$  \\
%     $\m_1 \oplus \m_2$ &  $\min(\m_1+\m_2,1)$ & $\max(\m_1,\m_2)$ & $\m_1+\m_2-\m_1\cdot \m_2$ & $\max(\m_1,\m_2)$  \\
%     $\m_1 \impf \m_2$ & $\min(1-\m_1+\m_2,1)$ &
%     $\begin{cases} 1 & \mbox{\!if } \m_1\leq \m_2  \\
%      d_2 & \mbox{\!otherwise}\end{cases}$
%      & $\min(1, \m_2/\m_1)$ & $\max(1-\m_1,\m_2)$  \\[2.5ex]
%     ${\ominus}\, \m$ & $1-\m$ &
%     $\begin{cases} 1 & \mbox{\!if } \m=0\\
%                                  0 & \mbox{\!otherwise}\end{cases}$
%      & $\begin{cases} 1 & \mbox{\!if } \m=0\\
%                                  0 & \mbox{\!otherwise}\end{cases}$ & $1-\m$\\ \hline
% \end{tabular}
%
\end{center}
}
%\vspace*{-1ex}
%\vspace*{-1ex}
%\vspace{-5ex}
\end{table}

%We will also use an optional subscript $X \in \{l,g,p\}$ to identify
%the logic they refer to (\eg~$\alpha \otimes_{g} \beta$ refers to
%G{\"o}del conjunction). 

% Note that the operators for `standard' fuzzy logic,
% namely  $\m_1 \otimes \m_2 = \min(\m_1,\m_2)$, $\m_1 \oplus
% \m_2 =\max(\m_1,\m_2)$, ${\ominus}\, \m = 1-\m$ and
% $\m_1 \impf \m_2 = \max(1-\m_1,\m_2)$, can be expressed in
% \L{}ukasiewicz logic. More precisely, $\min(\m_1,\m_2) = \m_1
% \otimes_l (\m_1 \Rightarrow_l \m_2), \max(\m_1,\m_2) = 1-
% \min(1-\m_1, 1-\m_2)$. Furthermore, the implication  $\m_1
% \impf_{kd} \m_2 = \max(1-\m_1,\m_2)$ is called \emph{Kleene-Dienes
% implication} (denoted $\Rightarrow_{kd}$), while \emph{Zadeh
% implication} (denoted $\Rightarrow_{z}$) is the implication $\m_1
% \Rightarrow_{z} \m_2   = 1$  if $\m_1 \leq \m_2$; $0$ otherwise.

%
%\[
%\alpha \Rightarrow_{z} \beta   =  \left\{
%    \begin{array}{ll}
%        1, & \textrm{ if } \alpha \leq \beta \\
%        0 & \textrm{ if } \alpha > \beta \ .
%        \end{array}
%        \right.
%\]

An \emph{r-implication} is an implication function obtained as the
\emph{residuum} of a continuous t-norm $\otimes$,
% \footnote{Note that {\L}ukasiewicz, G\"{o}del and Product implications are
% r-implications, while Kleene-Dienes implication is not.} 
\ie~$\m_1 \Rightarrow \m_2 = \sup\{\m_3\mid \m_1 \otimes \m_3
\leq \m_2 \}$.
%
%\[
%\alpha \Rightarrow \beta = \max\{\gamma\mid \alpha \otimes \gamma \leq \beta \} \ .
%\]
%
\nd Note also, that given an r-implication $\Rightarrow_{r}$, we may
also define its related negation $\ominus_{r} \m$ by means of $\m
\Rightarrow_{r} 0$ for every $\m \in \unit$.

%Table~\ref{tabpropnorm} shows some additional properties of truth
%combination functions.
%%
%\vspace*{-2ex}
%%
%\begin{table}
%\caption{Some additional properties of truth combination functions.}
%\label{tabpropnorm}
%%
%\vspace*{-2.5ex}
%%{\footnotesize
%{\scriptsize
%%{\tiny
%\[
%\begin{array}{cccc||c} \hline
%\text{Property} & \mbox{\L{}ukasiewicz logic} &  \mbox{G{\"{o}}del}  & \mbox{Product} & \mbox{Zadeh~\cite{Zadeh65}} \\ \hline
%\alpha \otimes {\ominus}\, \alpha  =  0  & \bullet & \bullet & \bullet & \\
%\alpha\oplus {\ominus}\, \alpha =  1  & \bullet  &   &   &  \\
%\alpha\otimes \alpha=  \alpha&  & \bullet &   &\bullet \\
%\alpha\oplus \alpha=  \alpha&  & \bullet &   &\bullet \\
%{\ominus}\, {\ominus}\, \alpha=  \alpha&\bullet &   &   &\bullet \\
%\alpha \impf \beta = {\ominus}\, \alpha \oplus \beta&\bullet &   &   &\bullet \\
%  \alpha \impf \beta = {\ominus} \beta \impf  \ominus \alpha & \bullet &   &   & \bullet \\
%{\ominus}\, (\alpha \impf \beta) = \alpha \otimes {\ominus}\, \beta &\bullet &   &   &\bullet \\
%{\ominus}\, (\alpha\otimes \beta) = {\ominus}\, \alpha \oplus {\ominus}\, \beta&\bullet  & \bullet & \bullet &\bullet \\
%{\ominus}\, (\alpha\oplus \beta) = {\ominus}\, \alpha \otimes {\ominus}\, \beta&\bullet  & \bullet & \bullet &\bullet \\  \hline \hline
%\alpha\otimes ( \beta \oplus \gamma) = (\alpha \otimes  \beta) \oplus  (\alpha \otimes  \gamma) &  &\bullet & &\bullet \\
%\alpha\oplus ( \beta \otimes \gamma) = (\alpha \oplus  \beta) \otimes  (\alpha \oplus  \gamma) &  &\bullet & &\bullet \\  \hline
%\end{array}
%\]
%}
%\vspace{-4ex}
%\end{table}

The notions of satisfiability and logical consequence are defined in
the standard way, where a fuzzy interpretation $\I$ \emph{satisfies}
a fuzzy formula $\fuzzyg{\phi}{\m}$, or  $\I$ is a \emph{model} of
$\fuzzyg{\phi}{\m}$, denoted as $\I\,{\models}\,\fuzzyg{\phi}{\m}$,
iff $\I(\phi) \geq \m$. Notably, from $\fuzzyg{\phi}{\m_1}$ and $\fuzzyg{\phi \rightarrow \psi}{\m_2}$ one may conclude (if $\rightarrow$ is interpreted as an r-implication)
$\fuzzyg{\psi}{\m_1 \otimes \m_2}$ (this inference is called \emph{fuzzy modus ponens}).

%\vspace{-2ex}
%--------
\paragraph{Fuzzy Description Logics basics.} \label{sect:fuzzyDL}
%\subsection{Fuzzy Description Logics basics} \label{sect:fuzzyDL}
%\vspace{1ex}
%\nd {\bf Fuzzy Description Logics basics.}

% \nd We recap here the fuzzy DL~$\alc(\D)$~\cite{Straccia05d,Bobillo11c,Straccia13}. $\alc(\D)$ is expressive enough to capture the main ingredients of fuzzy DLs we are going to consider here.

\nd We recap here the fuzzy DL~$\alcaggr(\D)$, which extends the well-known fuzzy DL $\alc(\D)$~\cite{Straccia05d} with the  \emph{aggregated concept} construct~\cite{Bobillo13a} (indicated with the symbol $\aggr$). $\alcaggr(\D)$ is expressive enough to capture the main ingredients of fuzzy DLs we are going to consider here.

% Note that fuzzy DLs and  fuzzy OWL 2 in particular,  cover many more language constructs than we use here (see, \eg~\cite{Bobillo15b,Bobillo11c,Straccia13}).

We start with the notion of \emph{fuzzy concrete domain}, that is a tuple
$\D\eqs\tuple{\Delta^{\D}, {\,\cdot\,}^{\D}}$ with datatype
domain~$\Delta^{\D}$ and a mapping ${\,\cdot\,}^{\D}$ that assigns
to each data value an element of $\Delta^{\D}$, and to every $1$-ary
datatype predicate $\mathbf{d}$ a $1$-ary fuzzy relation over $\Delta^{\D}$.
Therefore, ${\,\cdot\,}^{\D}$  maps indeed each datatype predicate
into a function from $\Delta^{\D}$ to $[0,1]$.
In the domain of numbers, typical datatypes predicates $\mathbf{d}$  are characterized by the well
known membership functions (see also Fig.~\ref{fig:muf})
%
%{\small
\begin{eqnarray*}
\mathbf{d} & \rightarrow &  ls(a,b)  \ | \ rs(a,b) \ | \   tri(a,b,c)  \ | \ trz(a,b,c,d)  \\
&& | \  \ \geq _{v} \ | \ \leq _{v} \ | \ = _{v}  \ ,
%\ | \ \\
%& & \ | \ \geq _{v} \ | \ \leq _{v} \ | \ = _{v}
\end{eqnarray*}
%}
\nd where additionally $\geq _{v}$ (resp.~$\leq _{v}$ and $=_{v}$) corresponds to the crisp set of data values that are no less than (resp.~no greater than and equal to) the value $v$.
%~\footnote{Specifically, $\geq _{v}$, $\leq _{v}$, and $= _{v}$ maybe seen as the same as $rs(v,v)$, $ls(v,v)$ and $tri(v,v,v)$, respectively)}

\emph{Aggregation Operators} (AOs) are mathematical functions that are used to combine different pieces of information. There exist  large number of different AOs that differ on the assumptions on the data (data types) and about the type of information that we can incorporate in the model~\cite{Torra07}.
There is no standard definition of AO. Usually, given a domain $\mathbb{D}$
(such as the reals), an AO of dimension $n$ is a mapping $\aggr:
\mathbb{D}^{n} \to \mathbb{D}$. For us, $\mathbb{D} = \unit$. Thus, an AO
aggregates $n$ values of $n$ different criteria. In our scenario,
such criteria will be represented by using fuzzy concepts from a
fuzzy ontology and we assume to have a finite  family  $\aggr_1, \ldots, \aggr_l$ of AOs within our language.

%\paragraph{Syntax.}
%

Now, consider pairwise disjoint alphabets ${\bf I}, {\bf A}$ and ${\bf R}$, 
where  ${\bf I}$ is the set of \emph{individuals}, 
${\bf A}$ is the set of \emph{concept names} (also called \emph{atomic concepts} or \emph{class names})
and ${\bf R}$ is the set of \emph{role names}. 
Each role is either an  \emph{object property} or  a \emph{datatype property}.
The set of \emph{concepts} are built from concept names $A$  using
connectives and quantification constructs over object properties $R$
and  datatype properties $S$, as described by the following
syntactic rule ($n_i\geq 1)$:
%, \alpha_i \in (0,1], \sum_i \alpha_i \leq 1$):
%
%{\small
\begin{eqnarray*}
C  & \rightarrow &  \topc \ | \ \bottomc \ | \ A\ |\ C_{1} \andc C_{2} \ |\
C_{1} \orc C_{2} \ |\ \notc C \ |\ C_{1} \rimp C_{2}  \ |\ 
\\
&& 
\some R.C \ | \ \all R.C  \ |\   \some S.\mathbf{d}  \ |\  \all S.\mathbf{d}   \ | \
\\
&& \aggr_i(C_1, \ldots, C_{n_i}) \ .
%D  & \rightarrow & \some T.\mathbf{d}  \ |\  \all T.\mathbf{d}  \ ,
\end{eqnarray*}
%}
%\vspace{-1ex}
%
%\nd Each of the connectives $\andc$ and $\orc$ may also have a
%subscript $X \in \{g, l, p\}$, $\rimp$ may  have a subscript $X \in
%\{kd, g, l, p, z \}$ and, $\notc$ may  have a subscript $Y \in \{g,
%l\}$. The subscript indicates the fuzzy logic the connectives refers
%to (see Section~\ref{sect:fuzzy-logic}). For instance, $ C \andc (D
%\rimp_{l} \all R. \notc_{g} E)$ is a concept (if a subscript is
%missing, then we assume that a priori selected fuzzy logic $X \in
%\{g, l, p, z \}$ has been selected).
%
An \emph{ABox} $\A$  consists of a finite set of assertion axioms.
An \emph{assertion} axiom is an expression of the form
\fuzzyg{\cass{a}{C}}{\m} (called \emph{concept assertion},  $a$ is an
instance of concept $C$ to degree greater than or equal to $\m$) or of the form
$\fuzzyg{\rass{a_{1}}{a_{2}}{R}}{\m}$ (called \emph{role assertion},
$(a_{1}, a_{2})$ is an instance of object property $R$ to degree greater than or equal to
$\m$), where $a, a_{1}, a_{2}$ are individual names, $C$ is a
concept, $R$ is an object property and $\m \in \unitp$ is a truth value.
A \emph{Terminological Box} or \emph{TBox} $\T$ is a finite set of
\emph{General Concept Inclusion} (GCI) axioms, where a fuzzy GCI is of the form
$\fuzzyg{C_{1} \impc C_{2}}{\m}$ ($C_{1}$ is a sub-concept of
$C_{2}$ to degree greater than or equal to $\m$), where $C_{i}$ is a concept and $\m
\in \unitp$. 
%
%A \emph{primitive} GCI is one of the form $\fuzzyg{A
%\impc C}{\m}$, where $A$ is a concept name and $C$ is a concept. In
%both cases above, $\impc$ may also have a subscript $X \in \{g, l,
%p, z \}$. Note that $\impc_{kd}$ is not allowed.
%A \emph{definitional} GCI is one of the form $A \defc C$. $A$ is
%called the \emph{head} of a primitive/definitional axiom, and $C$ is
%the \emph{body}.
%%
%A \emph{synonym} GCI is of the form $A \defc B$, where both $A$ and
%$B$ are concept names.
%%
%A \emph{generalised definitional} GCI is of the form $C \defc D$,
%where both $C$ and $D$ are concepts.
%
%A \emph{constraint} axiom has one of the following form ($R$ is an object property):
%\ii{i} $\dom(R,C)$, called \emph{domain restriction}, that restricts the
%domain of role $R$ to be concept $C$;
%\ii{ii} $\rg(R,C)$, called \emph{range restriction},  that restricts the
%range of role $R$ to be concept $C$; and
%\ii{iii} $\disj(A,B)$, called \emph{disjoint restriction},  that restricts
%the concept names $A$ and $B$ to be disjoint.
%
%
%\begin{itemize}
%%
%\item $\dom(R,C)$, called \emph{domain restriction}, that restricts the
%domain of role $R$ to be concept $C$;
%%
%\item $\rg(R,C)$, called \emph{range restriction},  that restricts the
%range of role $R$ to be concept $C$;
%%
%\item $\disj(A,B)$, called \emph{disjoint restriction},  that restricts
%the concept names $A$ and $B$ to be disjoint.
%%
%%\item $\bin(A)$, that restricts atomic concept $A$ to be bivalent;
%%\item $\bin(R)$, that restricts role $R$ to be bivalent.
%\end{itemize}
%
We may omit the truth degree $\m$ of an axiom; in this case $\m = 1$
is assumed and we call the axiom \emph{crisp}. We also write $C_1 = C_2$ as a macro for the two GCIs $C_1 \impc C_2$ and $C_2 \impc C_1$.
We may also call a fuzzy GCI of the form $\fuzzyg{C \impc A}{\m}$, where $A$ is a concept name, a \emph{rule} and $C$ its \emph{body}.
A \emph{Knowledge Base} (KB) is a pair $\K = \tuple{\T, \A}$, where $\T$ is a TBox and $\A$ is an ABox. With $\indkb$ we denote the set of individuals occurring  in $\KB$.

%Let $\T$ be a TBox consisting of definitional GCIs only. Concept
%name $A$ \emph{directly uses} concept name $B$ \wrt~$\mathcal{T}$,
%if $A$ is the head of some axiom $\tau \in \T$ such that $B$ occurs
%in the body of $\tau$. Let \emph{uses} be the transitive closure of
%the relation directly uses. $\T$ is \emph{acyclic} if no concept
%name $A$ is the head of more than one definitional axiom in $\T$ and
%there is no concept name $A$ such that $A$ uses $A$.

%\paragraph{Semantics.}
%
Concerning the semantics, let us fix a fuzzy logic, a fuzzy concrete domain $\D\eqs\tuple{\Delta^{\D}, {\,\cdot\,}^{\D}}$
and aggregation operators 
$\aggr_i: \unit^{n_i} \to \unit$.
Now, unlike classical DLs in which an interpretation $\I$
maps \eg~a concept $C$ into a set of individuals $\highi{C}
\subseteq \highi{\Delta}$,  \ie~$\I$ maps $C$ into a function
$\highi{C}: \highi{\Delta} \to \{0, 1\}$ (either an individual
belongs to the extension of $C$ or does not belong to it), in fuzzy
DLs, $\I$ maps $C$ into a function $\highi{C}: \highi{\Delta} \to
\unit$ and, thus, an individual belongs to the extension of $C$ to
some degree in $[0,1]$, \ie~$\highi{C}$ is a fuzzy set.
Specifically, a \emph{fuzzy interpretation} is a pair $\I =
(\highi{\Delta}, \highi{\cdot})$ consisting of a nonempty (crisp)
set $\highi {\Delta}$ (the \emph{domain}) and of a \emph{fuzzy
interpretation function\/} $\highi{\cdot}$ that assigns: \ii{i} to
each atomic concept $A$ a function $\highi{A}\colon\highi{\Delta}
\rightarrow \unit$; \ii{ii} to each object property $R$ a function
$\highi{R}\colon\highi{\Delta} \times \highi{\Delta} \rightarrow
\unit$; \ii{iii} to each datatype property $S$ a function
$\highi{S}\colon\highi{\Delta} \times \Delta^{\D} \rightarrow
\unit$; \ii{iv} to each individual $a$ an element $\highi{a} \in
\highi{\Delta}$ such that $\highi{a} \neq \highi{b}$ if $a \neq b$ (the so-called \emph{Unique Name Assumption}); and \ii{v} to each data value $v$ an element
$\highi{v} \in \Delta^{\D}$.
Now, a fuzzy interpretation function is extended to %roles and
concepts as specified below (where $x \in \highi{\Delta}$):
%
%\vspace*{-1ex}
{
%\small
%{\footnotesize
%{\scriptsize
%\[
%\begin{array}{l}
\begin{eqnarray*}
&&\highi{\topc}(x)  =   1 \  ,  \ \highi{\bottomc}(x)  \ =  \ 0 \  ,  \ 
\highi{(C \andc D)}(x)   =    \highi{C}(x) \andf \highi{D}(x)  \\
&&\highi{(C \orc D)}(x)   =    \highi{C}(x) \orf \highi{D}(x) \  ,  \ 
\highi{(\notc C)}(x)   =   \notf \highi{C}(x) \\
&&\highi{(C \rimp D)}(x)   =    \highi{C}(x) \impf \highi{D}(x)   \ ,  \ 
\highi{(\all R.C)}(x)    =    \inf_{y \in \highi{\Delta}} \{ \highi{R}(x,y) \impf \highi{C}(y) \} \\ 
&&\highi{(\csome R.C)}(x)   =    \sup_{y \in \highi{\Delta}} \{ \highi{R}(x,y) \andf \highi{C}(y) \} \  ,  \
\highi{(\all S.\mathbf{d})}(x)   =    \inf_{y \in  \Delta^{\D}} \{ \highi{S}(x,y) \impf \mathbf{d}^{\D}(y) \} \\ 
&&\highi{(\csome S.\mathbf{d})}(x)   =    \sup_{y \in  \Delta^{\D}} \{ \highi{S}(x,y) \andf \mathbf{d}^{\D}(y) \} \  ,  \\
% \highi{(\alpha_1 \cdot C_1 + \ldots \alpha_n \cdot C_n)}(x)   =    \sum_i \alpha_i \cdot \highi{C_i}(x) \ . 
&&\highi{(\aggr_i(C_1, \ldots, C_{n_i}))}(x)   = \aggr_i(\highi{C_1}(x), \ldots,  \highi{C_{n_i}}(x)) \ . 
\end{eqnarray*}
%\end{array}
%\vspace{-1ex}
%\]
}
%
%
%\nd Hence, for every  concept $C$ we get a function $C^{\I}: \highi{\Delta} \to \unit$.
%
\nd The \emph{satisfiability of axioms} is then defined by the
following conditions:
\ii{i} $\I$ satisfies an axiom $\fuzzyg{\cass{a}{C}}{\m}$ if    $C^{\I} (a^{\I}) \geq \m$;
\ii{ii} $\I$ satisfies an axiom $\fuzzyg{\rass{a}{b}{R}}{\m}$ if  $R^{\I} (a^{\I}, b^{\I}) \geq \m$;
%\ii{iii} $\I$ satisfies an axiom $\fuzzyg{\rass{a}{v}{T}}{\m}$ if  $T^{\I} (a^{\I}, v^{\I}) \geq \m$;
 \ii{iii} $\I$ satisfies an axiom $\fuzzyg{C \impc D}{\m}$ if $\highi{(C \impc
D)} \geq \m$ with\footnote{However, note that under standard logic
$\impc$ is interpreted as $\impf_{z}$ and not as $\impf_{kd}$. } $
  \highi{(C \impc D)} = \inf_{x \in \highi{\Delta}} \{ \highi{C}(x) \impf \highi{D}(x) \}
$.
$\I$ is  a \emph{model} of  $\KB = \tuple{\A, \T}$ iff $\I$ satisfies each
axiom in $\KB$. If $\KB$ has a model we say that $\KB$ is \emph{satisfiable} (or \emph{consistent}).
We say that $\KB$ \emph{entails}  axiom $\tau$,
denoted $\KB \models \tau$, if any model of $\KB$ satisfies
$\tau$. The \emph{best entailment degree} of $\tau$ of the form $C \impc D$, $\cass{a}{C}$ or $\rass{a}{b}{R}$, denoted $\bed{\KB}{\tau}$, is defined as
\[
\bed{\KB}{\tau} = \sup\{ \m \mid \KB \models \fuzzyg{\tau}{\m}\} \ .
\] 

\begin{remark} \label{remneg}
Please note that $\bed{\KB}{\cass{a}{C}} = 1$ (\ie~$\K \models \cass{a}{C})$ implies that $\bed{\KB}{\cass{a}{\notc C}} = 0$ holds, and similarly, 
$\bed{\KB}{\cass{a}{\notc C}} = 1$ (\ie~$\K \models \cass{a}{\notc C})$ implies that $\bed{\KB}{\cass{a}{C}} = 0$ holds. However, in both cases the other way around does not hold. 
Furthermore, we may well have that both $\bed{\KB}{\cass{a}{C}} = \m_1 > 0$ and $\bed{\KB}{\cass{a}{\notc C}} = \m_2 > 0$ hold.
\end{remark}
Now, consider concept $C$, a rule $C \impc A$, a KB $\KB$ and a set of individuals $\mathsf{I}$. 
% and a (weight) distribution $\myvec{w}$ over $\ind$. 
Then the \emph{cardinality} of $C$ \wrt~$\KB$ and $\mathsf{I}$, denoted $|C|_\KB^{\mathsf{I}}$, is defined as 
\begin{equation} \label{card}
|C|_\KB^{\ind} = \sum_{a \in \ind} \bed{\KB}{\cass{a}{C}} \ .
\end{equation}
% \nd while the \emph{weighted cardinality} $C$ \wrt~$\KB$, $\myvec{w}$ and $\ind$ , denoted $|C|_\KB^{\myvec{w},\ind}$, is defined as 
% \begin{equation} \label{wcard}
% |C|_\KB^{\myvec{w},\ind} = \sum_{a \in \ind} w_a \cdot \bed{\KB}{\cass{a}{C}} \ .
% \end{equation} 

% \nd The \emph{crisp cardinality} (denoted $\ceil{C}_\KB^{\ind}$) and \emph{crisp weighted cardinality}  (denoted $\ceil{C}_\KB^{\myvec{w},\ind}$) are defined similarly by replacing in Eq.~\ref{card} and \ref{wcard} the term $\bed{\KB}{\cass{a}{C}}$ with $\ceil{\bed{\KB}{\cass{a}{C}}}$.

\nd The \emph{crisp cardinality} (denoted $\ceil{C}_\KB^{\ind}$) is defined similarly by replacing in Eq.~\ref{card} the term $\bed{\KB}{\cass{a}{C}}$ with $\ceil{\bed{\KB}{\cass{a}{C}}}$.

Eventually, we say that the \emph{application} of rule $C \impc A$ to individual $a$ \wrt~$\KB$ is $\bed{\KB}{\cass{C}{a}}$ and that rule $C \impc A$ \emph{applies} to individual $a$ \wrt~$\KB$ if $\bed{\KB}{\cass{C}{a}} > 0$.

% Furthermore, the \emph{confidence degree} (also called  \emph{inclusion degree}) of $C \impc D$ \wrt~$\KB$ and $\mathsf{I}$, denoted $cf(C \impc D, \ind)$, is defined as
% \begin{equation}\label{cf}
% cf(C \impc D, \ind) =  \frac{|C \andc D |_\KB^{\ind}}{|C|_\KB^{\ind}} \ .
% \end{equation}

%\todo{support}
%\facomment{support mai definito. Ho reintegrato le linee che seguono}

% The \emph{support} of $C \impc D$ \wrt~$\KB$ and $\mathsf{I}$, denoted $supp(C \impc D, \KB, \ind)$, is defined as
% \begin{equation}\label{supp}
% supp(C \impc D, \KB, \ind) =  \frac{|C|_\KB^{\ind}}{|\ind|} \ 
% \end{equation}

% \nd instead.

% \nd Similarly,  the \emph{weighted confidence degree} (also called  \emph{weighted inclusion degree}) of $C \impc D$ \wrt~$\KB$, $\myvec{w}$ and $\mathsf{I}$, denoted $cf(C \impc D, \myvec{w}, \ind)$, is defined as
% \begin{equation}\label{wcf}
% cf(C \impc D, \myvec{w},\ind) =  \frac{|C \andc D |_\KB^{\myvec{w},\ind}}{|C|_\KB^{\myvec{w},\ind}} \ .
% \end{equation}

%%%%%%%%%%%%%%%%%%%%%%%%%%%%%%%%%%%%%%%%%%%%%%%%%%%%%%%%%%%%%%%%%%%%

\section{\fuzzyowlpnrule}
\label{sec:learn}

\nd At first, we introduce our learning problem.

%------
\subsection{The Learning Problem} \label{sect:tlp}
%------

\nd In general terms, the learning problem we are going to address is stated as follows. 
Consider 
\begin{enumerate}
\item a satisfiable crisp KB $\K$ and its individuals $\indkb$;
\item a \emph{target concept name} $T$;
\item an associated classification function  $f_T \colon \indkb \to \{-1,0,1\}$, where for each $a\in \indkb$,
 the values (\emph{labels}) correspond to 
 \begin{equation*}
  f_T(a) =
    \begin{cases}
      1 & \text{$a$ is a \emph{positive} example \wrt~$T$}\\
      -1 & \text{$a$ is a \emph{negative} example \wrt~$T$}\\
      0 & \text{$a$ is an \emph{unlabelled} example \wrt~$T$}
    \end{cases}       
\end{equation*}
%  \begin{enumerate}
%      \item $a$ is a \emph{positive} example of $T$, case $f_T(a)=1$; 
%      %$\K \models \cass{a}{T}$
%      \item $a$ is a \emph{negative} example of $T$, case $f_T(a)=-1$;
%       %$\K \models \cass{a}{\notc T}$
%      \item $a$ is an \emph{unlabelled} example of $T$, case $f_T(a)=0$;
%      %neither $\K \models \cass{a}{T}$ nor $\K \models \cass{a}{\notc T}$ holds
%  \end{enumerate}

\item the partitioning of the examples into
\begin{flalign*}
\posi  & =  \{(a,1) \mid a \in \indkb, f_T(a) = 1 \}  \text{ \Comment{the positive examples}}\\
\nega  & =  \{(a,-1) \mid a \in \indkb, f_T(a) = -1 \}  \text{ \Comment{the negative examples}} \\
\unlabel  & =  \{(a,0) \mid a \in \indkb, f_T(a) = 0 \}  \text{ \Comment{the unlabelled examples}} \\
\end{flalign*}

\nd where $\posi \neq \emptyset$ is assumed. We define $\calE = \posi \cup \nega \cup \unlabel$ as the set of all examples, and with $\overline{\posi} = \calE \setminus \posi$ we denote the set of \emph{non-positive} examples.

\item the set of individuals $\inds{\calS} = \{a \mid (a,l) \in \calS \}$, where $\calS \subseteq \calE$ is a set of examples. Moreover, we define 
\begin{flalign*}
\inds{\posi}  & =  \{a \mid (a,1) \in \posi \}  \text{ \Comment{the positive individuals}}\\
\inds{\nega}  & =  \{a \mid (a,-1) \mid a \in \nega \}  \text{ \Comment{the negative individuals}} \\
\inds{\unlabel}  & =  \{a \mid (a,0) \mid a \in \unlabel \}  \text{ \Comment{the unlabelled individuals}} \\
\inds{\overline{\posi}}  & =  \indkb \setminus \inds{\posi} \text{ \Comment{the non-positive individuals}} \\
\end{flalign*}

 \item  a \emph{hypothesis space} of classifiers 
 $\mathcal{H} = \{h \colon \indkb \to [0,1]\}$; 
 
 \item a \emph{training set} $\train \subset \calE$  of individual-label pairs, with $\train \cap \posi \neq \emptyset$;

\item a \emph{test set} $\test = \calE \setminus \train$.

% \item a set $\unlabel = \KB \setminus (\posi \cup  \nega)$, we denote with $\inds^{\unlabel}$ the set of \emph{unlabelled} individuals $\inds^{\unlabel}$ = $\indkb \setminus (\inds{\posi} \cup \inds{\nega})$.  

%disjoint from $\train$.
\end{enumerate}

% \nd Given a subset $\mathcal{E} \subseteq \posi \cup  \nega$, we denote with $\inds_{\mathcal{E}}$ the set of individuals occurring in $\mathcal{E}$ and define $\inds^{\unlabel}$ = $\indkb \setminus \inds{\mathcal{E}}$ as the set of \emph{unlabelled} individuals \wrt~$\mathcal{E}$.  

% We also call $a\in \inds{\posi}$ (resp.~$a\in \inds{\nega}$ and $a\in \inds{\unlabel}$) a \emph{positive} (resp.~\emph{negative} and \emph{unlabelled}) example.

% We also call $a\in \inds{\posi}$ (resp.~$a\in \inds{\nega}$ and $a\in \inds^\unlabel$) a \emph{positive} (resp.~\emph{negative} and \emph{unlabelled}) example.

\nd We assume that the only axioms involving $T$ in $\KB$ are either of the form $\cass{a}{T}$ or $\cass{a}{\neg T}$.
% for all $a\in \indkb$, $0 = \bed{\KB}{\cass{a}{T}} = \bed{\KB}{\cass{a}{\neg T}}$. That is we state that $\K$ does not already know whether $a$ is an instance of $T$ or of $\notc T$. 
We write $\E(a) =  1$ if $a$ is a positive example (\ie  $a \in  \inds{\posi}$), $\E(a) =  -1$ if $a$ is a negative example (\ie  $a \in  \inds{\nega}$) and $\E(a) =  0$ otherwise.

The general goal is to learn a classifier function $\bar{h} \in \mathcal{H}$ that is the result of \emph{Empirical Risk Minimisation} (ERM) on $\train$, \ie 
\begin{eqnarray*}
\bar{h} & = & \arg \min_{h \in \mathcal{H}} R(h, \train)  
%& = &  \mathbf{E}_{\mathcal{E}}[L(h(a),\E(a))] \\
\  = \ \frac{1}{|\train|} \sum_{a \in \inds{\train}} L(h(a),\train(a)) \ ,
\end{eqnarray*}
%
% \begin{align*}
% \bar{h} & =  \arg \min_{h \in \mathcal{H}} R(h, \mathcal{E})  \\
% \intertext{and}
% %& = &  \mathbf{E}_{\mathcal{E}}[L(h(a),\E(a))] \\
% R(h, \mathcal{E})  & = \ \frac{1}{|\mathcal{E}|} \sum_{a \in \inds{\E}} L(h(a),\E(a)) \ ,
% \end{align*}

\nd where $L$ is a \emph{loss function} such that $L(\hat{l}, l)$  measures how different the prediction $\hat{l}$ of a hypothesis is from the true label $l$ and $R(h,\train)$ is the \emph{risk} associated with hypothesis $h$ over $\train$, defined as the expectation of the loss function over the training set $\train$. 

The effectiveness of the learnt classifier $\bar{h}$ is then assessed by determining $R(\bar{h}, \test) $ on the test set $\test$. 

In our learning setting, a hypothesis $h \in \mathcal{H}$ is a set of GCIs of the form 
\begin{align}
 \fuzzyg{C_{1}  \impc   P_1}{\alpha_1} \ , \ldots \ , 
\fuzzyg{C_{h}  \impc   P_h}{\alpha_h}  \label{eqnhyp1}  \\ 
 \aggr^+(P_{1}, \ldots, P_h)  \impc   P  \label{eqnhyp2} \\
 \nonumber \\
 \fuzzyg{D_{1}  \impc   N_1}{\beta_1} \ , \ldots \ , 
\fuzzyg{D_{k}  \impc  N_k}{\beta_k} \label{eqnhyp3}  \\
\aggr^-(N_{1}, \ldots, N_k)  \impc   N  \label{eqnhyp4} \\
\nonumber \\
 \aggr(P,N)  \impc   T  \label{eqnhyp5}
\end{align}

%
% \begin{equation} \label{eqnhyp}
% \begin{array}{rcl}
% \alpha_1 \cdot T_1 + \ldots + \alpha_n \cdot T_n &  \impc  & T \\
% C_{ij} & \impc  & T_i  \ , \text{ with }1 \leq i \leq n, 1 \leq j \leq k_i \ ,
% \end{array}
% \end{equation}

\nd where each $P_i, P, N_j, N$ are  new atomic concept names not occurring in $\KB$, and $\alpha_i, \beta_j$ are  the confidence degree of the relative GCIs, $\aggr^+, \aggr^-, \aggr$ are aggregation operators, and each $C_{i}, D_j$ is a fuzzy $\EL(\D)$ concept expression defined as ($v$ is a boolean value)
\begin{equation*}\label{eq:left-hand-GCI}
\begin{array}{lcl}
C & \longrightarrow &  \top \mid A \mid \some r.C \mid \some s.\mathbf{d} \mid C_{1} \andc C_{2} \\
\dt & \rightarrow  & ls(a,b) \ | \  rs(a,b) \ | \ tri(a,b,c) \  | \  trz(a,b,c,d) \ | \  = _{v} \ .
\end{array}
\end{equation*}

\nd  Informally, \ii{i} each $P_i$ `rule' will tell us why an individual should be positive; \ii{ii} then we aggregate the various degrees of positiveness via the aggregator operator $\aggr^+$; \ii{iii} on the other hand, each $N_i$ `rule' will tell us why an individual should be \emph{not} positive; \ii{iv} then we aggregate the various degrees of non-positiveness via the aggregator operator 
$\aggr^-$. Typically, both $\aggr^+$ and $\aggr^-$ are the $\max$ operator; finally, \ii{v} we use the last `rule' to establish whether and individual is an instance of $T$ or not (\viz~is positive or not positive) by combining the degree of being positive or not via the $\aggr$ operator. A simple choice for $\aggr$ is the following and will be the one we will adopt: 

\begin{quote} \label{finalagg}
($\star$) if the degree $p$ of being positive is greater than the degree of being non-positive $n$ then $p$, else $0$.
%\footnote{See, Section~\ref{conceptex} for further clarification on how we build rules of Eqs.~(\ref{eqnhyp1})- (\ref{eqnhyp5}).} 
\end{quote}

% , which can be formalised, \eg~as
% $
% \aggr(p,n) = p \otimes (n \impf p)
% $
% under \mbox{G{\"{o}}del} logic.

% \nd  Essentially, each $\alpha_i$ indicates how well the `union' of the $C_{i1}, \ldots, C_{ik_i}$ contributes to classify an individual $a$ as being an instance of $T$. Specifically, if $\alpha_i > 0$ then $T_i$ contributes to $a$'s positiveness, while if $\alpha_i < 0$ then $T_i$ contributes to $a$'s non-positiveness instead.

% \begin{remark}\label{el-}
% Please note that we do not learn expressions of the form \eg~$\csome s.= _{v}$ for integer/real values $v$ as the search space would be too large and they would be likely non-effective. This is the reason why we restrict the $C_{ij}$ in Eq.~\ref{eqnhyp} to fuzzy $\elbool$ concept expressions and not fuzzy $\el(\D)$ instead.
% \end{remark}

\nd Now, for  $a\in \indkb$, the \emph{classification prediction value} $h(a)$ of $a$ , $T$ and $\K$ is defined as %(for ease, we omit $\K$ and $T$)
\begin{equation} \label{predval}
%h(a) = \bed{\KB \cup \{h\} }{\cass{a}{T}} \ .
h(a) = \bed{\KB \cup h }{\cass{a}{T}} \ .
\end{equation}

\begin{remark} \label{posnegrem}
\nd Note that, as stated above, essentially a hypothesis is a sufficient condition for being an individual instance of a target concept to  some degree. If $h(a) = 0$ then we say that $a$ is not a positive instance of $T$, while if $h(a) > 0$ then $a$ is a positive instance of $T$ to  degree $h(a)$. 
As a consequence, we will distinguish between positive and non-positive examples of $T$ only. That is, negative examples and unlabelled examples are indistinguishable. 
% Nevertheless, the framework may also distinguish
% between negative and non-negative examples of $T$ simply by considering a new target concept name, say  $Non\_T$, whose positive examples are the negative examples of $T$ and the negative examples are the positive ones of $T$ and then `learn'  $Non\_T$. 
\end{remark}

\nd Let us note that even if \K\ is a crisp KB, the possible occurrence of fuzzy concrete domains in expressions of the form $\some S.\mathbf{d}$ in a hypothesis may imply that not necessarily $h(a)  \in \{0,1\}$. A similar effect may also be induced by the aggregation operators. 

%\begin{remark}
%Note that in \eg~\cite{Lisi15} a hypothesis is of the form $\fuzzyg{C_1 \orc \ldots  \orc C_n \impc T}{\m}$ instead.\qed
%\end{remark}

\begin{remark}
Clearly,  the set of hypotheses by this syntax is potentially infinite due, \eg~to conjunction and the nesting of existential restrictions in the concept expressions. This set is made finite by imposing further restrictions on the generation process such as the maximal number of conjuncts and the maximal depth of existential nestings allowed.
\end{remark}

\nd We conclude by saying that a hypothesis $h$ \emph{covers} (resp.~$\theta$-covers,  for $\theta \in (0,1]$) an individual $a\in \indkb$ iff $h(a) > 0$ (resp.~$h(a) \geq \theta$), and indicate with $Cov(h)$ (resp.~$Cov_\theta(h)$) the set of covered (resp.~$\theta$-covered) individuals. Moreover, for a GCI 
$C \impc T$, the \emph{confidence degree} (also called  \emph{inclusion degree}) of $C \impc T$ \wrt~$\KB$ and a set of positive individuals $P$, denoted $cf(C \impc T, \KB, P)$, is defined as
\begin{equation}\label{cf}
cf(C \impc T, \KB, P) =  \frac{|C|_\KB^{P}}{|C|_\KB^{\indkb}} \ ,
\end{equation}

\nd which is the proportion of positive individuals covered by $C$ \wrt~the individuals covered by $C$. Clearly, $cf(C \impc T, \KB, P) \in [0,1]$ and the closer the confidence is to $1$ the 
`more precise' is $C \impc T$, in the sense the less it covers non-positive individuals. In addition, the \emph{support} of $C \impc T$ \wrt~$\KB$ and a set of individuals $\mathsf{I}$, denoted $supp(C \impc T, \KB, \ind)$, is defined as
\begin{equation}\label{supp}
supp(C \impc T, \KB, \ind) =  \frac{|C|_\KB^{\ind}}{|\ind|} \ 
\end{equation}

% Eventually, the \emph{mean} of $T$ \wrt~$\KB$ and a set of positive examples $P$, 
% $\mu(T, \KB, \ind)$, is defined as 
% \begin{equation}\label{meanT}
% \mu_T(\KB, P) =  \frac{|P|}{|\indkb|} \ .
% \end{equation}

% \nd Note that $0 < \mu_T(\KB, P) < 1$, as the the extreme values (there are no positive examples,  all individuals are positive examples, respectively) are not of interest in our learning setting.

% %\nd Note that in fact $\mu(T, \KB, \ind) = supp(T \impc T, \KB, \ind)$.

% We say that a hypothesis $h$ \emph{covers} (\emph{strongly covers}) an example $e \in \mathcal{E}$ iff $\bed{\KB \cup h}{e} > 0$ ($\bed{\KB \cup h}{e} = 1$). Therefore, soundness states that a  learnt hypothesis is not allowed to cover a non-positive example, while the way (strong) completeness is stated  guarantees that all positive examples are (strongly) covered.

% In general a learnt (\emph{induced}) hypothesis $h$ has to be \emph{consistent}, \emph{non-reduntant} and  \emph{sound} \wrt~$\K$, but not necessarily complete, but, of course, these conditions can also be relaxed.

%%%%%%%%%%%%%%%%%%%%%%%%%%%%%%%%%%%%%%%%%%%%%%%%%%%%%%%%%%%%%%%%%%%%

%---------------------
%\subsection{The Learning Algorithm \fuzzyowlpnrule} \label{sect:tlpa}
%---------------------

%------
\subsection{Conceptual Illustration of the Learning Method.}  \label{conceptex}
%\paragraph{Conceptual Illustration of the Learning Method.}
\begin{figure}[t]
%\begin{scriptsize}
\begin{center}
\begin{tabular}{c@{\ \ \ \ \ \ \ \ }c@{\ \ \ \ \ \ \ \ }c@{\ \ \ \ \ \ \ \ }c}
\includegraphics[scale=0.25]{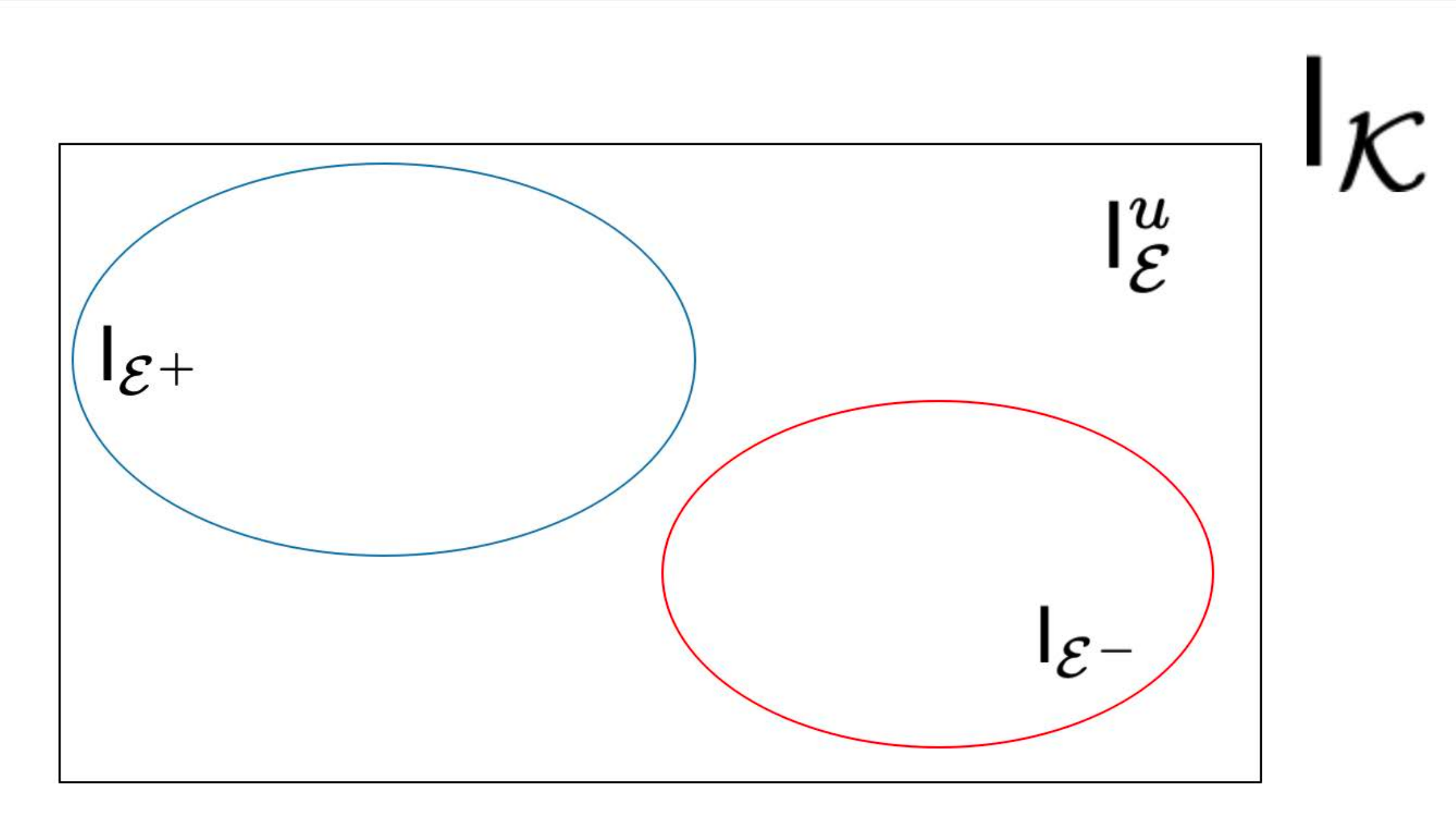} &
\includegraphics[scale=0.25]{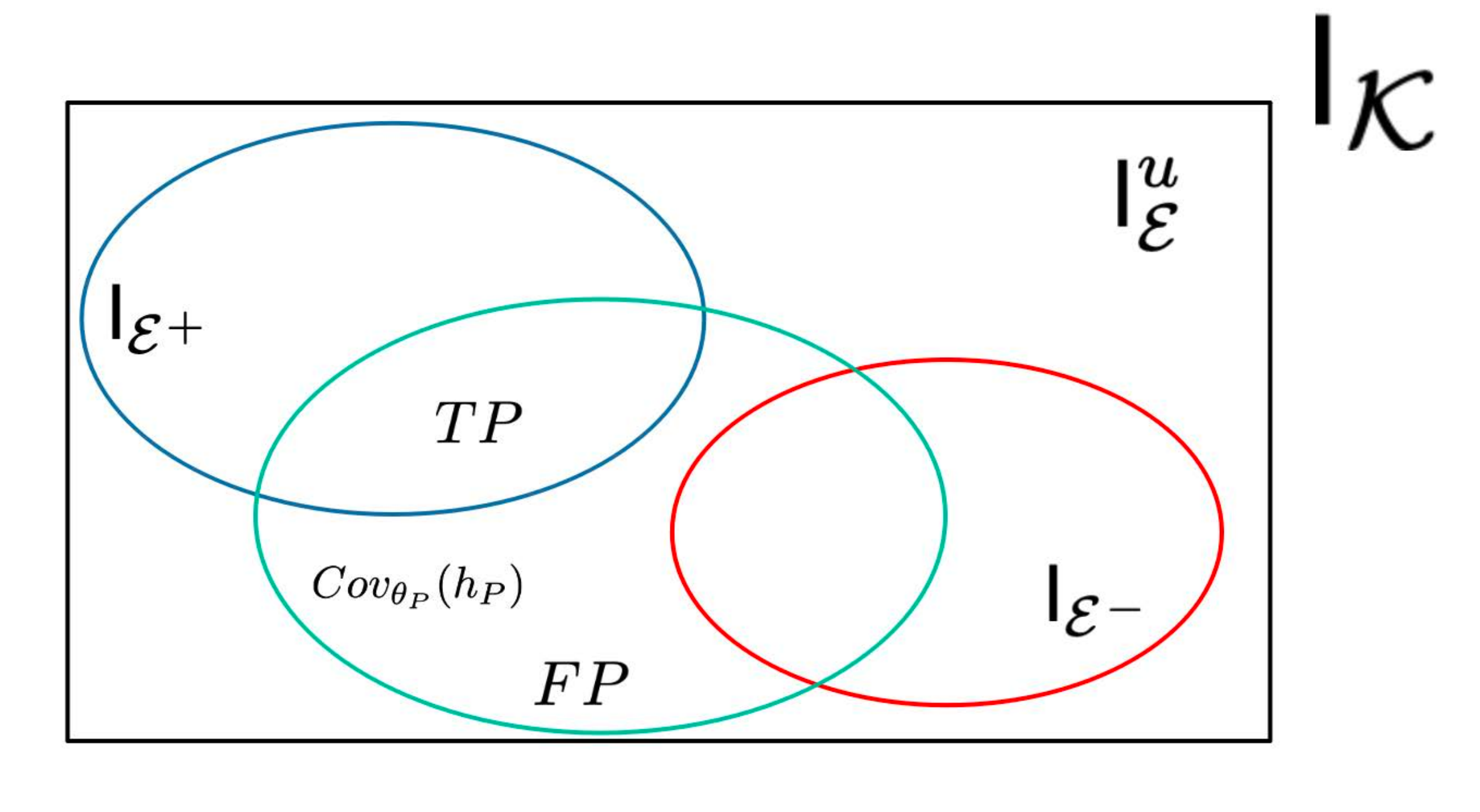} & \\
(a) & (b)   \\
\includegraphics[scale=0.25]{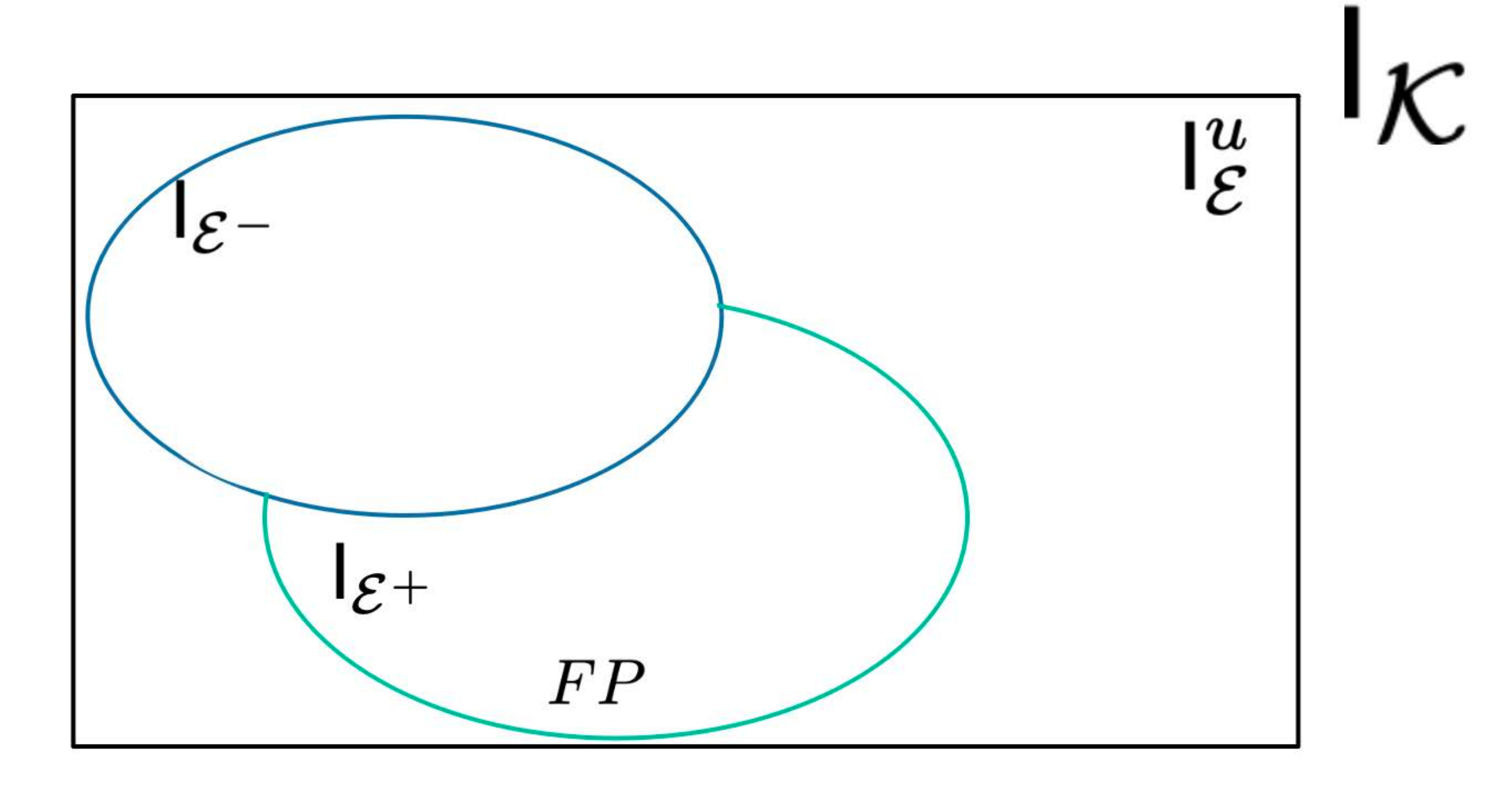} &
\includegraphics[scale=0.25]{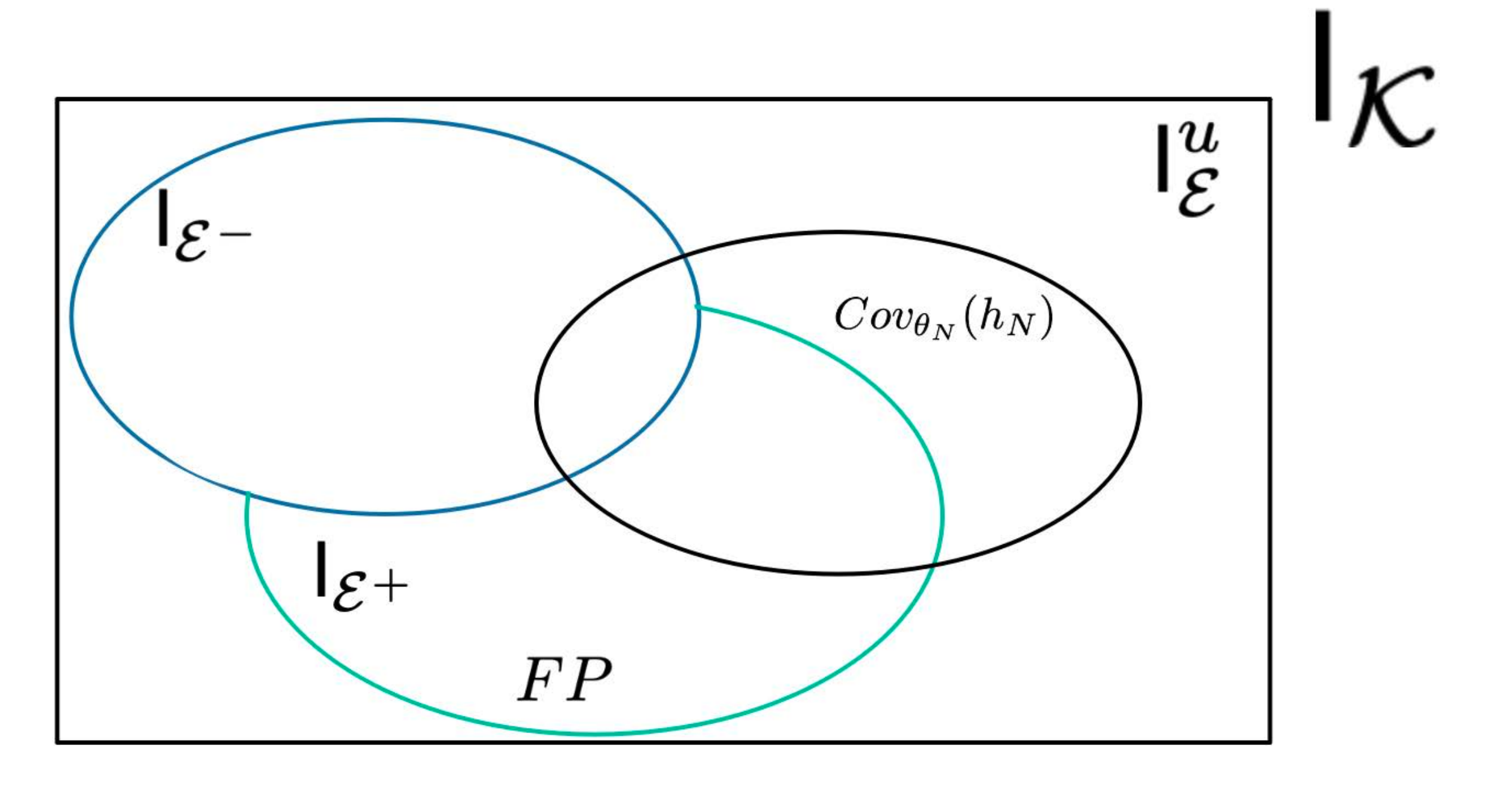} \\
(c) & (d) \\
\includegraphics[scale=0.25]{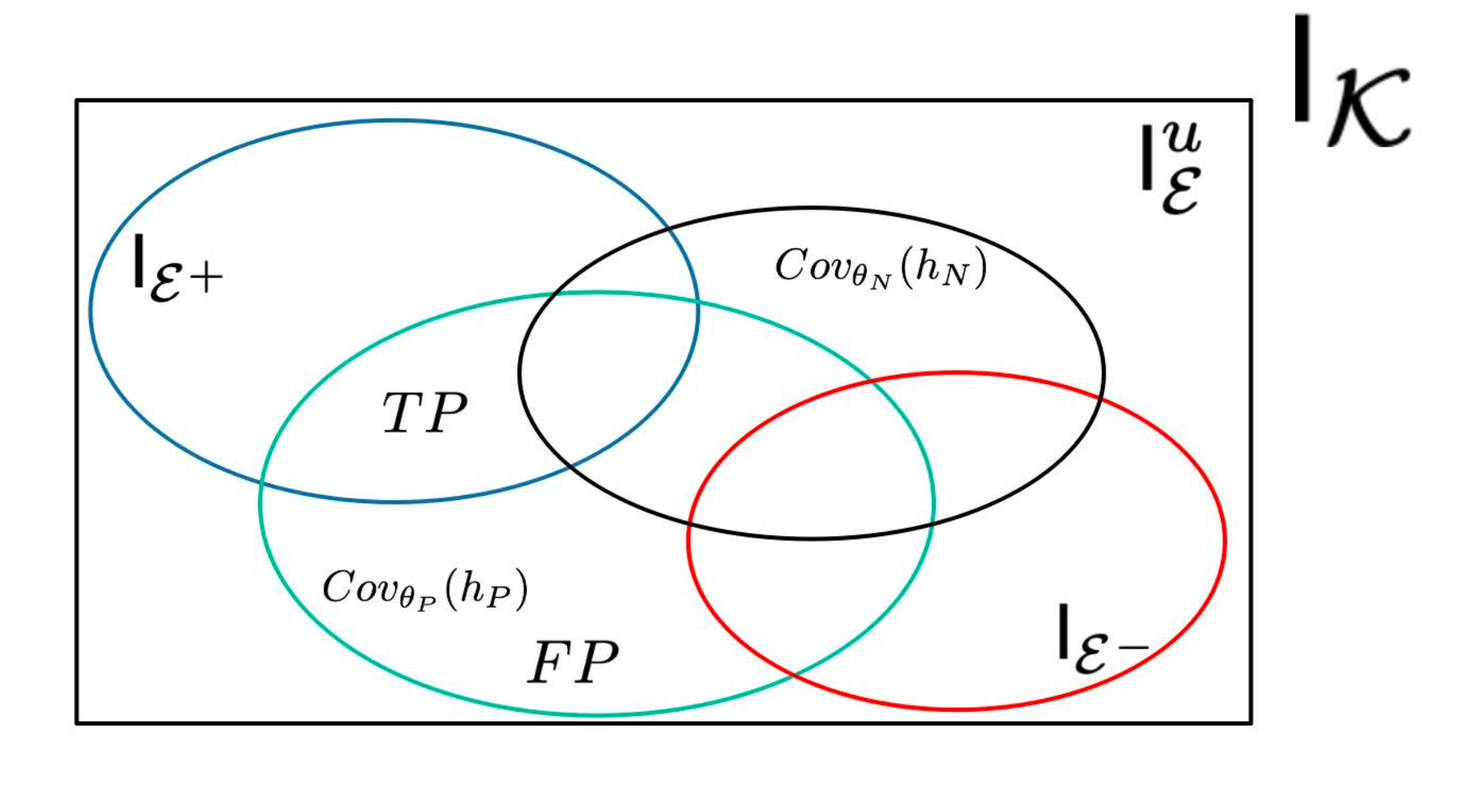} & \\
(e) & 
%(e & (f)
\end{tabular}
%\vspace*{-1ex}
\caption{How \fuzzyowlpnrule~works. (a) Original training set;
(b) Coverage $Cov_{\theta_P}(h_P)$ \wrt~learnt hypothesis $h_P$ after the P-stage; (c) Starting dataset for N-stage: the new target class are the false positives $FP$ of the P-stage, while the negative individuals are the initial positives; (d) Coverage $Cov_{\theta_N}(h_N)$ \wrt~learnt hypothesis $h_N$ after the N-stage; (e) Final scenario.}
\label{pnconcept}
%\vspace*{-5ex}
%\vspace{-.5cm}
\end{center}
%\end{scriptsize}
\end{figure}
%\vspace{-5ex}
Before presenting our learning algorithm, we will first conceptually illustrate its principle by relying on Figure~\ref{pnconcept}.

At the beginning, let us consider the sets of all individuals, the positive, the negative and the unlabelled individuals, respectively the sets $\indkb, \inds{\E^+},  \inds{\E^-}$ and 
$\inds{\unlabel}$, as depicted in Figure~\ref{pnconcept} (a). 

At the first stage, the P-stage, we consider the entire training set $\E$ and try to maximise the covering of positive individuals, while minimising the covering of negative individuals. Specifically, let us assume that we have learnt a hypothesis $h_P$ (a set of rules)  with a covering $Cov_{\theta_P}(h_P)$, as depicted in Figure~\ref{pnconcept} (b).
Here, the value $\theta_P$ acts as a confidence threshold for the learnt rules in hypothesis $h_P$. Note that $Cov_{\theta_P}(h_P)$ has to contain positive individuals, \ie$Cov_{\theta_P}(h_P) \cap \inds{\E^+} \neq \emptyset$, but may also contain negative and unlabelled individuals. We call the individuals in 
$TP = Cov_{\theta_P}(h_P) \cap \inds{\E^+}$ \emph{true positives}, while call those in
$FP = Cov_{\theta_P}(h_P) \setminus \inds{\E^+}$ \emph{false positives}, \ie a false positive is an individual that is erroneously classified by $h_P$ as an instance of the target class $T$, while in fact it is not (it might be an unlabelled or a negative example). This phase ends with a set of rules of the form $(\ref{eqnhyp1})-(\ref{eqnhyp2})$.

Now, in the next stage, the N-stage, with the aim to increase the effectiveness of the classifiers, we would like to remove as many as possible false positives in $FP$, while avoiding removing, if possible, any of the true positives in $TP$. To do so, we set-up a new learning problem in which the new target class is $FP$, where the negatives individuals are those in $TP$ and the positives are those in $FP$. Of course, the N-stage applies only if $FP \neq \emptyset$. The setup of the N-stage is depicted in Figure~\ref{pnconcept} (c). Specifically, let us assume that we have learnt now a hypothesis $h_N$ with a covering $Cov_{\theta_N}(h_N)$, as depicted in Figure~\ref{pnconcept} (d). Note that we may have another parameter $\theta_N$ acting as a confidence threshold for the learnt rules in hypothesis $h_N$. 
This phase ends with a set of rules of the form $(\ref{eqnhyp3})-(\ref{eqnhyp4})$.

So, in general, at the end of the two stages, the situation may be as depicted in Figure~\ref{pnconcept} (e).
However, in practice one may want likely to impose that none of the initial positive individuals are covered by $h_N$ and, thus, none of the true positives in $TP$ will be removed by $h_N$.
%, and being  $\bar{\theta}$ as high as possible.

Eventually, we aggregate the P-rules and N-rules via ($\star$). This latter step ends with the rule of the form (\ref{eqnhyp5}). At the end of this two-stage process, we aim at to have captured most of the positive individuals of the target class, with few of the negative and unlabelled individuals (false positives).
%\vspace{-4ex}

%\mycomment{
% Most of the false positives still getting covered can be attributed to the lower confidence of P-rules. Similarly, most of the positive examples missing from the coverage can be attributed to the lower confidence N-rules.
% Based on this observation, we design a scoring mechanism, \ie~a suitably aggregation operator of $h_P$ with $h_N$,
% that allows to recover as positive some of the false positives introduced by the low confidence N-rules. 
% This latter step ends with the rule of the form $(\ref{eqnhyp5})$.

% Also, the scoring mechanism will try to assign low scores to the false positives examples covered by low confidence P-rules. Note that we can afford to be more aggressive by keeping the final confidence threshold low in each of the stages, because we rely on our scoring mechanism to correct for the additional errors introduced. 

%---------------------
\subsection{The Learning Algorithm \fuzzyowlpnrule} \label{sect:tlpa}
%\paragraph{The Learning Algorithm.}
%---------------------

\nd We now present our two-stage learning algorithm, called \fuzzyowlpnrule, that we have conceptually illustrated in the section before. 
Essentially, at the P-stage (resp.~N-stage) our algorithm invokes a learner, called \emph{stage learner}, that generates a set $h_{P}$ (resp.~$h_{N}$) of fuzzy $\eld$ candidate GCIs that has, respectively, the form 
\begin{eqnarray}
h_P & = & \{ \fuzzyg{C_{1} \impc T}{\alpha_{1}}, \ldots, \fuzzyg{C_{h} \impc T}{\alpha_{h}} \} \label{hp} \\
h_N & = & \{ \fuzzyg{D_{1} \impc FP}{\beta_{1}}, \ldots, \fuzzyg{D_{k} \impc FP}{\beta_k} \} \label{hn}
\end{eqnarray}

\nd called  \emph{stage hypothesis}. 
In the following, we indicate with $p_i$ the fuzzy GCI  $\fuzzyg{C_{i} \impc T}{\alpha_i}$, while denote with $n_j$ the fuzzy GCI $\fuzzyg{D_j \impc FP}{\beta_j}$. The rules in $h_P$ (resp.~$h_N$) will then be aggregated using the $\max$ aggregation operator.

The stage hypotheses are then  combined into a final hypothesis for the target class $T$ using the aggregation operator $(\star)$.

As stage learner we will use a modified version of the fuzzy \foildl~\cite{Lisi13,Lisi13a,Lisi15} learner that will be described in Section~\ref{sect:wla}.

Then, the \fuzzyowlpnrule~algorithm is shown in \refalgo{alg:owlpnrule}. Note that the P-stage are the steps 1-5, while the N-stage are the steps 15-19 in which at step 19 we invoke the stage learner trying to cover as many as false positives as possible. The remaining steps deal with the construction of the final classifier ensemble as per Eqs.~(\ref{eqnhyp1})-(\ref{eqnhyp5}).

Eventually, for an individual  $a\in \indkb$, the \emph{classification prediction value of \fuzzyowlpnrule}~for individual $a$ is $h(a)$, where $h$ is the returned hypothesis of \fuzzyowlpnrule. Moreover, we say that $\fuzzyowlpnrule$ \emph{classifies} $a$ as instance of target class $T$ if $h(a) > 0$.

\begin{algorithm}[H]
%{\small
%{\footnotesize
\begin{algorithmic}[1]
%\FUNCTION{learnSetOfAxioms}{$\KB$, $T$, $\E^{+}$, $\E^{-}$, $\theta$, $\beta_1$, $\beta_2$}
\Require KB $\K$, training set $\E$, target concept name $T$, confidence thresholds $\theta_P, \theta_N \in [0,1]$, 
non-positive coverage percentages $\eta_P, \eta_N \in [0,1]$
\Ensure Hypothesis $h$ as by  Eqs.~(\ref{eqnhyp1})-(\ref{eqnhyp5}).
%Eq.~\ref{hypoex2a}.
%\State $h \gets \emptyset$;
%\State $\ind \gets \indkb$;
%\State $\calH' \gets \emptyset$;
%\State $\myvec{w}_1 \gets \myvec{u}$; \Comment{Initialize the weight distribution over $\ind$} 
%\State $h \gets \emptyset$;
\State // P-stage
\State $Pos \gets \inds{\E^+}$; 
\State $Neg \gets \inds{\E^-}$;
\State $U \gets \inds{\KB} \setminus (Pos \cup Neg)$;
%\For{$i=1$ to $n$} 
%\State $h_i \gets \fuzzyg{C_i \impc T}{\m_i}   \gets$ \Call{FuzzyWeakLearner}{$\KB$, $T$, $\E$, $\myvec{w}_i$}; \Comment{Weak learner $h_i$ is $\fuzzyg{C_i \impc T}{\m_i}$}
\State $h_P \gets$ \Call{FuzzyStageLearner}{$\KB$, $T$, $Pos$, $Neg$, $U$, $\theta_P$, $\eta_P$}; \Comment{P-Stage hypothesis $h_P$, \ie~set of axioms $\fuzzyg{C_{i} \impc T}{\alpha_i}$}
\If{$h_P = \emptyset$} \Return $\emptyset$; \Comment{Nothing learnt, exit} \EndIf
\State $Cov \gets Cov_{\theta_P}(h_P)$; \Comment{P-stage Coverage}
\State $TP \gets Cov_{\theta_P}(h_P) \cap \inds{\E^+}$; \Comment{True positives}
\State $FP \gets Cov_{\theta_P}(h_P) \setminus \inds{\E^+}$; \Comment{False positives}

\State // Start building classifier $h$
\State $h \gets \{\fuzzyg{C_{i} \impc P_i}{\alpha_{i}}, \mid  \fuzzyg{C_{i} \impc T}{\alpha_{i}} \in h_P, P_i \mbox{ new }\}$;
\Comment{As per Eq.~\ref{eqnhyp1}}

\If{$FP = \emptyset$} 
\Comment{No N-stage, exit with aggregated $h_P$} 
\State $h \gets h \cup \{ \aggr^+(P_{1}, \ldots, P_h)  \impc T \}$; \Comment{No need of new $P$ in Eq.~\ref{eqnhyp2}}
\State \Return $h$;
\EndIf

% \If{$FP = \emptyset$} {\bf GoTo} Step 16; \Comment{No N-stage} \EndIf

\State // N-stage
\State $Pos \gets FP$; 
\State $Neg \gets \inds{\E^+}$;
\State $U \gets \inds{\KB} \setminus (Pos \cup Neg)$;
\State $h_N \gets$ \Call{FuzzyStageLearner}{$\KB$, $FP$, $Pos$, $Neg$, $U$, $\theta_N$, $\eta_N$}; \Comment{N-Stage hypothesis $h_N$, \ie~set of axioms $\fuzzyg{D_{j} \impc FP}{\beta_j}$}
%  \State // Compute statistics
%  %\State $n_P \gets |h_P|$; \Comment{Number of rules in $h_P$}
%  %\State $n_N \gets |h_N|$; \Comment{Number of rules in $h_N$}
%  \State $ScoreMatrix \gets$ \Call{ComputeScore}{$suppm$, $errm$, $MinSuppScore$, $MinZ$ }; 
%  \State // ...
\State // Build final classifier ensemble $h$
\If{$h_N = \emptyset$} \Comment{No learning in N-stage, return aggregated $h_P$} 
\State $h \gets h \cup \{ \aggr^+(P_{1}, \ldots, P_h)  \impc T \}$; \Comment{No need of new $P$ in Eq.~\ref{eqnhyp2}}
\State \Return $h$;
\EndIf
\State $h \gets h \cup \{ \aggr^+(P_{1}, \ldots, P_h)  \impc P \mid P \mbox{ new } \}$; \Comment{As per Eq.~\ref{eqnhyp2}}
%\State // Build classifier ensemble as per Eqs.~(\ref{eqnhyp3})-(\ref{eqnhyp4}).

\State $h \gets h \cup \{\fuzzyg{D_{j} \impc N_j}{\beta_{j}}, \mid  \fuzzyg{D_{j} \impc FP}{\beta_{j}} \in h_N, N_j \mbox{ new }\}$;
\Comment{As per Eq.~\ref{eqnhyp3}}
\State $h \gets h \cup \{ \aggr^-(N_{1}, \ldots, N_k)  \impc N \mid N \mbox{ new } \}$; \Comment{As per Eq.~\ref{eqnhyp4}}
\State $h \gets h \cup \{ \aggr(P,N)  \impc   T \}$; \Comment{As per Eq.~\ref{eqnhyp5}}

\State \Return $h$;

%\ENDFUNCTION
\end{algorithmic}
%}
\caption{\fuzzyowlpnrule} \label{alg:owlpnrule}
\end{algorithm}

%---------------------
\subsection{The Stage Learner \foildlpn} \label{sect:wla}
%\paragraph{The Base Learner Fuzzy \foildl .} \label{sect:wla}
%---------------------

As stage learner we will use fuzzy \foildl~\cite{Cardillo22,Lisi13,Lisi13a,Lisi15}, which however will be modified to adapt to our specific setting (see  Algorithm~\ref{alg:wfoilsets}), which we call \foildlpn. That is, the procedure invocations {\sc FuzzyStageLearner} in lines 5 and 19 of the \fuzzyowlpnrule~algorithm are indeed calls to \foildlpn.

Essentially, \foildlpn~carries on inducing GCIs until as many as positive  examples are covered or nothing new can be learnt. When an axiom is induced (see step 4 in Algorithm~\ref{alg:wfoilsets}), the positive examples still to be covered are updated (steps 10 and 11).

%\mycomment{As for PN-rule, evaluate the case in which we remove all covered examples and not only the positive ones}

In order to induce an axiom (step 4),  \textsc{Learn-One-Axiom} is invoked (see Algorithm~\ref{alg:pnfoilOne}), which in general  terms operates as follows:
\begin{enumerate}
\item start from concept $\topc$;
\item apply a refinement operator to find more specific fuzzy $\eld$ concept description candidates;
\item exploit a scoring function to choose the best candidate;
\item re-apply the refinement operator until a good candidate is found; 
\item iterate the whole procedure until a satisfactory coverage of the positive examples is achieved.
\end{enumerate}
\begin{algorithm}[H]
\begin{algorithmic}[1]
\Require KB $\K$, target concept name $T$, a set $P$ (resp.~$N$ and $U$) of positive (resp.~negative and unlabelled) examples, confidence threshold $\theta \in [0,1]$, non-positive coverage percentage $\eta \in [0,1]$
%\Ensure Weak hypothesis of the form $\fuzzyg{C \impc T}{\m}$
%\Ensure A fuzzy $\elbool$ weak hypothesis of the form $C \impc T$
\Ensure A hypothesis, \ie~a set $h = \{ \fuzzyg{C_{i} \impc T}{\delta_{i}} | 1\leq i \leq k \}$ of fuzzy $\eld$ GCIs
%$h_i = \{ C_{i1} \impc T, \ldots, C_{ik_i} \impc T \}$ of fuzzy $\elbool$ GCIs.

%	\State $\theta\gets \mbox{init with a predefined value in } [0,1]$;
	%\State $h \gets \emptyset, Pos \gets P, \phi \gets \topc \impc T$; % NPos \gets \calE \setminus \calE^+$;
	\State $h \gets \emptyset, Pos \gets P,  \phi \gets \topc \impc T$; % NPos \gets \calE \setminus \calE^+$;
	%\ind \gets \indkb,
	%\State $\D \gets \textsc{InitialiseFuzzyConcreteDomain}(\K)$
	\State //Loop until no improvement
	\While{($Pos \neq \emptyset$) {\bf and} ($\phi \neq  \mathbf{null}$)} 
		\State $\phi \gets \Call{Learn-One-Axiom}{\KB, T, Pos, P, N, U, \theta, \eta}$; \Comment{ Learn one fuzzy $\eld$ GCI of the form $C \impc T$} % covering at least one positive example in $Pos$
        %\State $\fuzzyg{\phi}{d} \gets \Call{Learn-One-Axiom}{\KB, T, Pos, P, N, U, \theta, \eta}$; // Learn one fuzzy
        %$\eld$ GCI of the form $\fuzzyg{C \impc T}{d}$ % covering at least one positive example in $Pos$
		
		\If{$\phi \in h$} \Comment{axiom already learnt}
% 		\If{there is $\fuzzyg{\phi}{d'} \in h$ with $d'\geq d$} // better axiom already learnt
			\State $\phi \gets \mathbf{null}$; 
		\EndIf	
		\If{$\phi \neq  \mathbf{null}$}
			%\State $d \gets cf(\phi, \indkb, \KB\cup \{\phi\})$; // Compute confidence degree of $\phi$
% 			\State $\KB_\phi \gets \KB\cup \{\phi\}$; 
            %  \State $d \gets \frac{|C|_{\KB}^{P}}{|C|_{\KB}^{\indkb}}$; 
            %  // Compute confidence  of $\phi$
             \State $\delta \gets cf(\phi, \KB, P)$; 
             \Comment{Compute confidence of $\phi$}
% 			
             
% 			\State $d \gets \frac{|T|_{\KB\cup \{\phi\})}^{\indkb}}{|C|_{\KB\cup \{\phi\})}^{\indkb}}$; // Compute confidence degree of $\phi$
			\State $h \gets h \cup \{\fuzzyg{\phi}{\delta}\}$; \Comment{Update hypothesis}
			%\State $P_\phi \gets \{ \tuple{a,1} \in \calE^{+} \mid \bed{\K \cup \{\phi\}}{T(a) > 0} \}$; // Positives covered by $\phi$
			\State $Pos_\phi \gets Pos \cap Cov(\fuzzyg{\phi}{\delta})$; \Comment{Positives covered by $\fuzzyg{\phi}{\delta})$}
			
			\State $Pos \gets Pos \setminus Pos_\phi $; \Comment{Update positives still to be covered}
		\EndIf	
	\EndWhile
%\State $d \gets cf(\phi,\ind)$;	\Comment{Compute the  weak classifier  confidence degree}
%\State \Return $\fuzzyg{\phi}{d}$;
\State \Return $h$;
%\ENDFUNCTION
\end{algorithmic}
%\caption{\foildlw}
\caption{\foildlpn}
\label{alg:wfoilsets}
\end{algorithm}
\nd We now detail the steps of \textsc{Learn-One-Axiom} (Algorithm~\ref{alg:pnfoilOne}).

%----------------------------------
\vspace{1ex}
\nd {\bf Computing fuzzy datatypes.}
%\subsubsection{Computing fuzzy datatypes}
%---------------------------------
\nd For a numerical datatype $s$, we consider \emph{equal width triangular partitions} of values $V_s = \{ v \mid   \KB \models \cass{a}{\some s.=_{v}} \}$  into a finite number of fuzzy sets ($3,5$ or $7$ sets), which is identical to~\cite{Lisi13,Lisi15,Straccia15}  (see, \eg~Fig.~\ref{partfuzzytrz}). We additionally also consider the use of the c-means fuzzy clustering algorithm over $V_s$, where the fuzzy membership function is a triangular function build around the centroid of a cluster~\cite{Cardillo22,Lisi13,Lisi15,Straccia15}.

%----------------------------------
\vspace{1ex}
\nd {\bf The refinement operator.}
%---------------------------------
The refinement operator we employ is essentially the same as in~\cite{Cardillo22,Lisi13,Lisi13a,Lisi13b,Straccia15}. Specifically, it takes as input a concept $C$ and generates new, more specific concept description candidates $D$ (\ie, $\KB \models D \impc C$). For the sake of completeness, we recap the refinement operator here. Let $\KB$ be a knowledge base, ${\bf A}_\KB$ be the set of all atomic concepts in $\KB$, ${\bf R}_\KB$ the set of all object properties in $\KB$, ${\bf S}_\KB$ the set of all numeric datatype properties in $\KB$, ${\bf B}_\KB$ the set of all boolean datatype properties in $\KB$  and $\mathcal{D}$ a set of (fuzzy) datatypes. The refinement operator $\rho$ is shown in Table~\ref{tab:refinement}. 
\begin{table*}
\caption{Downward Refinement Operator.} \label{tab:refinement}
%\footnotesize
\small
\[
\rho(C)=\left\{
\begin{array}{lcl}
{\bf A}_\KB \cup \{ \some r.\topc \; | \; r \in {\bf R}_\KB \} \cup \{ \some s.d \; | \; s \in {\bf S}_\KB, d\in \mathcal{D} \}  \cup \\
\hspace{1cm}\{ \some s.=_b,\; | \; s \in {\bf B}_\KB, b \in \{ \mathbf{true},   \mathbf{false}\}\} 
& \mbox{if} & C=\topc \\

\{ A' \; | \; A'\in {\bf A}_\KB, \KB \models A' \impc A \} \cup \\
 \hspace{1cm} \{ A\andc A'' \; | \; A''\in\rho(\topc)\}
&  \mbox{if}  & C=A\\

\{ \some r.D'\; | \;D'\in\rho(D) \} \cup \{ 
(\some r.D) \andc D'' \; | \; D''\in \rho(\topc) \}
&  \mbox{if}  & C=\some r.D, r\in{\bf R}_\KB\\

\{ (\some s.d)\andc D \; | \; D\in \rho(\topc) \}
&  \mbox{if}  & C=\some s.d, s\in{\bf S}_\KB,d\in\mathcal{D}\\

\{ (\some s.=_b) \andc D \; | \; D\in \rho(\topc) \}
&  \mbox{if}  & C=\some s.=_b, s\in{\bf B}_\KB, \\
&& b\in\{\mathbf{true,false}\}\\

\{ C_1 \andc ... \andc C_i ' \andc ... \andc C_n \; | \; i=1,...,n,C_i '\in \rho(C_i)\}
&  \mbox{if}  & C=C_1 \andc ... \andc C_n\\
\end{array}
\right.
\]
\end{table*}

%----------------------------------
\vspace{1ex}
\nd {\bf The scoring function.}
%---------------------------------
The scoring function we use to assign a score to each candidate hypothesis is essentially a \emph{gain} function, like to the one employed in~\cite{Cardillo22,Lisi13,Lisi13a,Lisi13b,Straccia15}, and it implements an information-theoretic criterion for selecting the best candidate at each refinement step.
Specifically, given a fuzzy $\eld$ GCI $\phi$ of the form $C \impc T$ chosen at the previous step, a KB  $\KB$, a set of positive examples $Pos$ still to be covered and a candidate fuzzy $\eld$ GCI $\phi^\prime$ of the form $C^\prime \impc T$, then
\begin{equation}\label{eq:gain}
	gain(\phi^{\prime}, \phi, \KB, Pos) = p \ast (log_2(cf(\phi^{\prime},\KB, Pos)) - log_2(cf(\phi,\KB, Pos))) \ , 
\end{equation}

\nd where $p = |C^\prime \andc C |_\KB^{Pos}$ is the fuzzy cardinality of positive examples in $Pos$ covered by $\phi$ that are still covered by $\phi^{\prime}$. 
% and 
% \begin{equation}\label{wcfpos}
% cf(D \impc T, \myvec{w},\ind, Pos) =  \frac{|D|_\KB^{\myvec{w},\ind_{Pos}}}{|D|_\KB^{\myvec{w},\ind}} \ .
% \end{equation}

%\nd  Please note the change in Eq.~\refeq{wcfpos}~\wrt~Eq.~\refeq{wcf} concerning the confidence degree of $D \impc T$: in the former case, the numerator considers positive examples in $Pos$ only, \ie~the positive examples still to be covered (all instances of $T$ that are in $Pos$), while in the latter case all positives in $\calE^+$ (all instances of $T$) are considered. In this way,  \textsc{Learn-One-Axiom} is somewhat guided towards positives not yet covered  by the weak learner learnt so far by \foildlw.

Please note that in Eq.~\refeq{eq:gain}, the confidence degrees are calculated 
\wrt~the positive examples still to be covered ($Pos$). In this way,  \textsc{Learn-One-Axiom} is somewhat guided towards positives not yet covered so far by \foildlpn.
% At the beginning of \foildlw~we will have $Pos = \calE^+$, while at the end we may have  $Pos = \emptyset$.
Note also that the gain is positive if the confidence degree increases.
%----------------------------------

\vspace{1ex}
\nd {\bf Stop criterion.}
%---------------------------------
 \textsc{Learn-One-Axiom}~stops when the confidence degree is above a given threshold $\theta \in [0,1]$ and the non-positive coverage percentage is below 
 $\eta \in [0,1]$, or no GCI can be learnt anymore.

%----------------------------------
\vspace{1ex}
%\nd {\bf The \foildlw~Algorithm.}
\nd {\bf The \textsc{Learn-One-Axiom}~algorithm.}
%---------------------------------
The \textsc{Learn-One-Axiom}~algorithm just like defined in \refalgo{alg:pnfoilOne}: steps 1 - 3 are simple initialisation steps. Please note here that $NP$ are the non-positives in accordance with Remark~\ref{posnegrem}, which states that we will distinguish among positives and non-positives only (\cf~also step.~18, where the non-positive coverage percentage is used). 
Steps 5-21 are the main loop from which we may exit in case the stopping criterion is satisfied, in step 8 we determine all new refinements, which then are scored in steps 10-15 in order to determine the one with the best gain. At the end of the algorithm, once we exit from the main loop, the best found GCI is returned (step 22).

\begin{remark}
\foildlpn~also allows to use a backtracking mechanism (step 19), which, for ease of presentation,  we omit to include. The mechanism is the same as for the \pfoildl-learnOneAxiom described in~\cite[Algorithm 3]{Straccia15}. Essentially, a stack of  top-$k$ refinements is maintained, ranked in decreasing order of the confidence degree from which we pop the next best refinement (if the stack is not empty) in case no improvement has occurred. $C_{best}$ becomes the popped-up refinement.
\end{remark}
\begin{algorithm}[H]
\begin{algorithmic}[1]
%\Call{Learn-One-Axiom}{\KB, T, Pos, P, N, U, \theta, \eta}
\Require KB $\K$, target concept name $T$, set $Pos$ of positive examples still to be covered, training sets $P, N, U$ of positive, negative and unlabelled examples, respectively, confidence threshold $\theta \in [0,1]$, non-positive coverage percentage $\eta \in [0,1]$
%\Ensure Weak hypothesis of the form $\fuzzyg{C \impc T}{\m}$
\Ensure A fuzzy $\eld$ GCI of the form $C \impc T$
%	\State $\theta\gets \mbox{init with a predefined value in } [0,1]$;
% 	\State $\ind \gets \indkb, NP \gets N \cup U$;
	\State $NP \gets N \cup U$; \Comment{Note: $NP$ are the non-positives}
	\State $C \gets \topc$;  \Comment{Start from $\top$}
	\State $\phi \gets C \impc T$;
	\State //Loop until no improvement
	\While{$C \neq \mathbf{null}$} 
		\State $C_{best} \gets C$;
		%\State $\phi_{best} \gets C_{best} \impc T$;
		\State $maxgain\gets 0$;
		\State $\calC \gets \rho(C)$;  \Comment{Compute all refinements of $C$}
		\State // Compute the score of the refinements and select the best one
		\ForAll{$C^\prime \in\calC$}
			\State $\phi^\prime \gets C^\prime \impc T$;
			\State $gain\gets gain(\phi^{\prime}, \phi, \KB, Pos) $;
% 		    \If{$(gain > maxgain)$ \textbf{and} $(cf(\phi',\KB, Pos) >  cf(\phi,\KB, Pos))$}
			\If{$(gain > maxgain)$ }
				\State $maxgain\gets gain$;
				\State $C_{best}\gets C'$;
				%\State $C_{best}\gets C'$;
			\EndIf
		\EndFor
		%\If{$C_{best}=C$} \textbf{break}; \Comment{No improvement}
		\If{$C_{best}=C$}  \Comment{No improvement}
			\State //Stop if confidence degree above threshold or non-positive  coverage below threshold
%			\If{$(cf(C_{best} \impc T,\ind) \geq \theta)$ \textbf{and} $\frac{|C_{best}|_{\KB}^{\inds{\E^{-}}}}{|\inds{\E^{-}}|} \leq \eta$}   \textbf{break}; 
			\If{$(cf(C_{best} \impc T,\KB, P) \geq \theta)$ \textbf{and} %$(\frac{\ceil{C_{best}}_{\KB}^{NP}}{|NP|} \leq \eta)$}  
			$supp(C_{best} \impc T, \KB, NP)  \leq \eta)$}  
			\textbf{break}; 		
                 \EndIf{}
                 \State // Manage backtrack here, if foreseen
                 \EndIf{}
		\State $C\gets C_{best}$;
		\State $\phi \gets C \impc T$;
	\EndWhile
%\State $d \gets cf(\phi,\ind)$;	\Comment{Compute the  weak classifier  confidence degree}
%\State \Return $\fuzzyg{\phi}{d}$;
\State \Return $\phi$;
%\ENDFUNCTION
\end{algorithmic}
%\caption{\foildlw}
%\caption{\foildlw:learnOneAxiom}
\caption{\textsc{Learn-One-Axiom}}
\label{alg:pnfoilOne}
\end{algorithm}

%---------------------
\section{Evaluation}
\label{sec:eval}
%---------------------

\nd We have implemented the algorithm within the 
\emph{FuzzyDL-Learner}\footnote{Data and implementation \url{http://www.umbertostraccia.it/cs/software/FuzzyDL-Learner/}.} system and have evaluated it over a set of (crisp) OWL ontologies. 

%----------------------------------
\vspace{1ex}
\nd {\bf Datasets.}
%\subsection{Setup}
%----------------------------------
Several OWL ontologies from different domains have been selected as illustrated in Table~\ref{tab:onto}. In it, we report the DL the ontology refers to, the number of concept/class names, object properties, datatype properties and individuals in the ontology. For each ontology $\KB$ we indicate also the number $|\posi|$ of positive examples. All others are non-positive and we set  $\nega = \overline{\posi} =  \indkb \setminus \inds{\posi}$.
\begin{table*}
\caption{Facts about the ontologies of the evaluation.}
\label{tab:onto}
%\vspace*{-3ex}
%{\footnotesize
{\scriptsize
%{\tiny
%{\small
\begin{center}{
\begin{tabular}{l||cccccccc} \hline  \hline
\bf{ontology}  & \bf{DL} & \bf{class.}   & \bf{obj. prop.}   & \bf{data. prop.} & \bf{ind.} & \bf{target} $T$ & \bf{pos}    \\ \hline\hline
%FamilyTree & $\mathcal{SROIF(D)}$ & 22   & 52   & 6 & 368  & Uncle  & 46 & 156  & 1/5/0  \\ \hline 
%Hotel & $\mathcal{ALCOF(D)}$    & 89   & 3 & 1 & 88 & Good\_Hotel   & 12 & 11 & 1/5/0  \\ \hline 
%Moral & $\mathcal{ALC}$ & 46   & 0   & 0 & 202  & ToLearn\_Guilty & 102 & 100 &1/5/0  \\ \hline 
{\tt NTN} & $\mathcal{SHOIN(D)}$ & 51   & 29   & 9 & 723 & {\tt ToLearn\_Woman} & 46    \\ \hline 
%UBA & $\mathcal{SHI(D)}$ &  44   &  26   & 8 & 1268 & Good\_Researcher  & 22 & 113 & 1/5/0  \\ \hline 
%WineOnto &  $\mathcal{SHI(D)}$  & 178   & 15   &  7 & 138  & ToLearn\_DryWine  & 15 & - &1/5/0  \\ \hline 
%Pair50 &  $\mathcal{ALC}$  & 3   & 6   & 0 & 311  & ToLearn  &  20 & 29 &2/5/0  \\ \hline 
%Straight &  $\mathcal{ALC}$  & 3   & 6   & 0 & 347  & ToLearn  &  4 & 50 & 3/5/1.0  \\ \hline \hline
%Carcinogenesis &  $\mathcal{ALC(D)}$  & 143   & 4   &  15 & 22372  & ToLearn  &  162 & 136 & 1/5/0  \\ \hline 
%Hepatitis &  $\mathcal{ALC(D)}$  & 15   & 5   &  12 & 6812  & ToLearn  &  206 & 294 &1/5/0  \\ \hline 
{\tt Lymphography} &  $\mathcal{ALC}$  & 50   & 0   &  0 & 148  & {\tt ToLearn}  &  81   \\ \hline
{\tt Mammographic} &  $\mathcal{ALC(D)}$  & 20   & 3   &  2 & 975  & {\tt ToLearn}  &  445    \\ \hline 
%Mutagenesis &  $\mathcal{ALC(D)}$  & 87   & 5   &  6 & 14187  & ToLearn  &  13 & 29 &3/5/100  \\ \hline
%NCTRER &  $\mathcal{ALCI(D)}$  & 38   & 9   &  50 & 10209  & ToLearn  &  131 & - &1/5/0  \\ \hline 
%Pyrimidine &  $\mathcal{ALC(D)}$  & 2   & 0   &  27 & 74  & ToLearn  &  20 & 20 &1/5/1.0  \\ \hline 
%Suramin &  $\mathcal{ALC(D)}$  & 47   & 3   &  1 & 2979  & ToLearn  &  7   \\ \hline 
{\tt Malware} &  $\mathcal{ALH(D)}$  & 192   & 6   &  10 & 5669  & {\tt malware}  &  500   \\ \hline  
%VAD &  $\mathcal{AL(D)}$  & 10   & 0   &  4 & 36  & Amusing...  & 4 & 1/3  & 0.5\\ \hline 
{\tt Iris} &  $\mathcal{ALEHF(D)}$  & 4   & 0   &  5 & 150  & \begin{tabular}{c}
%Iris-setosa \\
{\tt Iris-versicolor} \\
{\tt Iris-virginica} \\
\end{tabular}  &  
\begin{tabular}{c}
% 51\\
50 \\
 50 \\
\end{tabular}    \\ \hline 
{\tt Wine} &  $\mathcal{ALEHF(D)}$  & 3   & 0   &  13 & 178  & \begin{tabular}{c}
%Iris-setosa \\
{\tt 1} \\
{\tt 2} \\
{\tt 3} \\
\end{tabular}  &  
\begin{tabular}{c}
59\\
71 \\
48 \\
\end{tabular}    \\ \hline 
{\tt Wine Quality} &  $\mathcal{ALEHF(D)}$  & 7   & 0   &  11 & 6497  & {\tt GoodRedWine}  &  217   \\ \hline 
{\tt Yeast} &  $\mathcal{ALEHF(D)}$  & 11   & 0   &  8 & 1462  & {\tt CYT}  &  444   \\ \hline \hline
\end{tabular}}
\end{center}
}
%\vspace*{-1ex}
%\vspace*{-1ex}
%\vspace{-5ex}
\end{table*}
The ontologies {\tt Iris}, {\tt Wine}, {\tt Wine Quality} and {\tt Yeast} are built from the well-known \emph{UC Irvine Machine Learning Repository} (UCIMLR)~\cite{Dua:2019} and have been transformed from the CSV format, provided by that repository, into OWL ontologies according to the procedure described in~\cite{Cardillo22}. In the {\tt Wine Quality} ontology, the {\tt quality} attribute has been removed as the positive examples (the {\tt GoodRedWine}s) are those having ``quality" greater than or equal to 7.

All other ontologies, except {\tt malware}, belong to the well-known SML-Bench dataset~\cite{Westphal19}.\footnote{See also, \url{https://github.com/SmartDataAnalytics/SML-Bench}} 
The  {\tt malware} ontology has been described in~\cite{svec21,svec23}.

For completeness, in Appendix~\ref{ontodesc}, a succinct description of what the ontologies are about is provided.

\begin{remark}
While evaluating ontology-based learning algorithms is untypical on  numerical datatype properties,\footnote{To the best of our knowledge, we are unaware of any evaluation of ontology-based methods on those data sets.} we believe it is interesting to do so as an important ingredient of our algorithm is the use of fuzzy concrete datatype properties to improve the human understandability of the classification decision process.
\end{remark}

\begin{remark} \label{remh}
We leave it for future work to look at \eg~methods to learn from the training data a threshold
$0 \leq \tau_p \leq 1$ such that $h$ predicts individual $a$ to be a positive example if $h(a) > \tau_p$. However, in this paper, we will always have $\tau_p=0$. 

More generally, unlike we do now, if we would like to distinguish the negative examples from the unlabelled ones, 
we may well learn a classifier $h^-$ for negative examples and then define a decision method that predicts an individual $a$ to be a positive (resp.~negative) example based on the prediction value $h(a)$ (resp.~$h^-(a)$) of $a$  being a \emph{positive} (resp.~negative) example. That is, depending on the pair $\tuple{h(a), h^-(a)}$, one may then define e decision criteria whether $a$ is a positive or negative example, or just leave the prediction as \emph{unknown} if there is not enough evidence of being one of the two.

% We leave it for future work to look at \eg~methods to learn from the training data thresholds 
% $0 \leq \tau_n < \tau_p \leq 1$ such that $h$ predicts $a$ to be a positive example if $h(a) \geq \tau_p$,
% $h$ predicts $a$ to be a negative example if $h(a) \leq \tau_n$, while $h$'s prediction for $a$ remains \emph{unknown} otherwise. So, in our setting, we will always have $\tau_n=0$ and, thus, the \emph{unknown} prediction does not occur, in accordance with the fact that all individuals are assumed either positive examples or non-positive examples.
\end{remark}

\nd{\bf Measures.} We considered the following effectiveness measures (see also~\cite{Straccia15,Cardillo22}), which, for the sake of completeness, we recap here. 
Specifically, consider a learnt classifier $h$ and let us assume to have added it to the KB $\K$. In our setting, we always have the condition that if the classifier prediction value $h(a)$ of an individual $a$ is non-zero then the learner classifies $a$ as an instance of $T$, \ie $h$ predicts $a$ to be a positive example iff $h(a)>0$.

In line with what we have said above, as all individuals are either positive or non-positive, we will consider the following measures, all of which are based on crisp cardinality (see also Eq.~\ref{card}).

\begin{description}
\item[True Positives:] denoted $TP$, is defined as the number of instances of $T$ that are positive 
\begin{equation} \label{fTP}
TP =  \ceil{T}_{\KB}^{\inds{\E^+}} 
\end{equation}

\item[False Positives:] denoted $FP$, is defined as  the number of instances of $T$ that are not positive 
\begin{equation} \label{fFP}
FP =   \ceil{T}_{\KB}^{\inds{\overline{\E^+}}} 
\end{equation}

% \item[Fuzzy True Non-Positive:] denoted $TNP_f$, is defined as  
% \begin{equation} \label{fTNP}
% %TNP_f =  |\inds{\E^-}| - FP_f\ ,
% TNP_f =  |\inds{\E^-}| - FP\ ,
% \end{equation}

% \item[Fuzzy False Non-Positive:] denoted $FNP_f$, is defined as  
% \begin{equation} \label{fFNP}
% %FNP_f =  |\inds{\E^+}| - TP_f\ ,
% FNP_f =  |\inds{\E^+}| - TP\ ,
% \end{equation}

\item[Precision/Confidence:] denoted $P$, is defined as  the fraction of true positives \wrt~the covered examples of $h$
\begin{equation} \label{prec}
%P_f =  \frac{|D \andc T |_{\KB}^{\indkb}}{|D|_\KB^{\indkb}} = cf(D \impc T, \indkb) = d \ ,
%P_f =   \frac{TP_f}{|D|_\KB^{\inds{\E}}} = cf(D \impc T, \inds{\E}) = d \ ,
P =   \frac{TP}{ \ceil{T}_\KB^{\inds{\E}}} 
\end{equation}

\item[Recall:] denoted $R$, is defined as  fraction of true positives \wrt~all positives
\begin{equation} \label{rec}
R =  \frac{TP}{|\inds{\E^+}|} \ ,
\end{equation}

\item[$F1$-score:] denoted $F1$, is defined as
\begin{equation} \label{f1eq}
F1 = 2 \cdot \frac{P\cdot R}{P + R} \ .
\end{equation}

%\item[Fuzzy Accuracy:] denoted $Acc_f$, is defined as
%\begin{equation*}
%Acc_f = \frac{TP_f + TNP_f}{|\inds{\E}|} \ , 
%\end{equation*}

% \item[Mean Squared Error:] denoted $MSE$, is defined as  
% \begin{equation}
% MSE = \frac{1}{|\inds{\E}|} \cdot \sum_{a \in \inds{\E}} (h(a) - l(a))^2 \ .
% \end{equation}
%\nd where  $h(a) \in [0,1]$ is the  \emph{classification prediction value} of $a$ \wrt~$h$, $T$,  which is defined as 
%\[
%h(a) = \bed{\KB \cup h}{\cass{a}{T}} \ .
%\]
%$\E(a) =  1$ if $a$ is a positive example (\ie, $a \in  \inds{\E^+}$), $\E(a) =  0$ if $a$ is a non-positive example (\ie, $a \in  \inds{\E^-}$), and 
%
\end{description}

\nd For each parameter configuration, a stratified $k$-fold cross validation design\footnote{Stratification means here that each fold contains roughly the same proportions of positive and non-positive instances of the target class.} was adopted (specifically, $k=5$) to determine the macro average of the above described performance measures. In all tests, we have that $\inds{\E} = \indkb$ and that, of course, there is at least one positive example in each fold. For each fold, during the training phase, we remove all assertions involving test examples from the ontology, and, thus, restrict the training phase to training examples only. 
% \begin{remark} \label{remove}
%     Removing test examples and the related assertions from the ontology during the training phase may easily be overlooked in  ontology-based machine learning. For instance, consider individuals $a$ and $b$, where $a$ is a positive training example, while $b$ is a test example. Assume further that $hasFriend(a,b)$ belongs to a KB $\KB$. Now, if $b$ and, thus, $hasFriend(a,b)$ is not removed during the training phase, then one may erroneously induce that the positive examples are those that `have a friend', which may be not the case otherwise.
% \end{remark}

All configuration parameters for the best runs are available from the downloadable data, which we do not report here. Some of the salient parameters, used within our algorithm, are reported in Table~\ref{tab:param}.

\begin{table}[H]
\caption{Some salient parameters of the \fuzzyowlpnrule~algorithm.}
\label{tab:param}
\begin{center}
\begin{tabular}{|c|l|}\hline\hline
     $\theta_P$ & confidence threshold for positive rules of P-stage \\\hline
     $\theta_N$ & confidence threshold for negative rules of N-stage \\\hline
     $\eta_P$ & non-positive coverage percentage threshold for positive rules of P-stage \\\hline
     $\eta_N$ & non-positive coverage percentage threshold for negative rules of N-stage \\\hline
     $c_P$ & maximal number of conjuncts for positive rules of P-stage \\\hline
     $c_N$ & maximal number of conjuncts for negative rules of P-stage \\\hline
     $d_P$ & maximal role depth for positive rules of P-stage \\\hline
     $d_N$ & maximal role depth for negative rules of P-stage \\\hline
     \hline\hline
\end{tabular}
\end{center}
\end{table}

A typical parameter setup is as follows, but may vary depending on the ontology and may be subject of a search for the optimal setting.
\begin{description}
    \item[P-stage.] $c_P=5, d_P=1, \theta_P = 0.1, \eta_P=1.0$
    \item[N-stage.] $c_N=10, d_N=1, \theta_N = 0.3, \eta_N=0.2$
\end{description}

\nd Let us briefly comment them. During the P-stage, we would like to increase recall, that is the percentage of covered positives \wrt~all positives. To this end, we choose a low positive rule confidence threshold $\theta_P$ and high non-positive coverage percentage threshold $\eta_P$. In the N-stage however, we want to be more precise in removing the false positives in order to avoid removing true positives of the P-Stage. Therefore, we increase the confidence threshold $\theta_N$, lower the non-positive coverage percentage threshold $\eta_N$ and increase the number of maximal conjuncts $c_N$. The maximal role depth is determined manually a priori by inspecting the ontology.

For $\aggr^+, \aggr^-$ (resp.~$\aggr$) we used $\max$ (resp.~($\star$)), and for concept conjunction $\andc$ (resp.~GCI operator $\impc$) we used the t-norm $\min$ (resp.~the {\L}ukasiewicz implication). These could well be another set of parameters to be optimised. However, the parameter space is already quite large, so we fixed these logical operators as specified.\footnote{A run with fixed parameters, \eg~on the {\tt malware} ontology, may already take up to 4 days of computation time.} Concerning other parameter settings, we also varied the number of fuzzy sets ($3, 5$ or $7$). For c-means, we fixed the hyper-parameter to the default $m=2$, the threshold to $\epsilon = 0.05$ and the number of maximum iterations to $100$. 
%Eventually, we impose that each learnt rule covers more than one positive example.

As baseline, we consider Fuzzy \foildl~\cite{Lisi13,Lisi13a,Lisi15,Cardillo22}, with best parameter setup as specified in~\cite{Cardillo22}. Essentially, Fuzzy \foildl, is as \fuzzyowlpnrule, except that it stops after the P-stage and, thus, is as \fuzzyowlpnrule~in which the negative set of rules $h_N$ is by definition empty (\cf~lines 21-23 of \fuzzyowlpnrule~algorithm). This allows us to appreciate the added value (if any) in terms of effectiveness of the N-stage phase. 

The results are reported in Table~\ref{tab:allres}. For the UCIMLR datasets, in case of multiple targets, the average of the measures has been considered.

\begin{example} \label{exh}
\nd We provide here examples of learnt rules  (in Fuzzy OWL syntax)  via \fuzzyowlpnrule~applied to the {\tt Mammographic} ontology. The first one is one of the learnt rules during the P-stage, while the second one is one of the learnt rules during the N-Stage. In the latter case, {\tt FALSEP\_ToLearn} denotes the class of false positives covered by rules learnt during the P-stage. The number associated to a rule is its confidence/precision. We also report the specification of some learnt fuzzy sets via fuzzy c-means.

%{\tiny
{\scriptsize
\begin{verbatim}
(implies (and (some hasDensity low)  
              (some hasShape irregular) 
              (some hasAge hasAge_veryHigh) 
              (some hasBiRads hasBiRads_high)) 
   ToLearn	0.965068)

(implies (and (some hasDensity low)  
              (hasMargin some microlobulated) 
              (hasShape some oval) 
              (hasBiRads some hasBiRads_medium))
   FALSEP_ToLearn	0.75)
   
(define-fuzzy-concept hasBiRads_medium    triangular(1,6,2.780,3.997,5.022))
(define-fuzzy-concept hasBiRads_high      right-shoulder(1,6,3.997,5.022))
(define-fuzzy-concept hasAge_veryHigh     right-shoulder(1,6,62.793,71.882))
\end{verbatim}
}

% %{\tiny
% {\scriptsize
% \begin{verbatim}
% (hasDensity some low) and (hasShape some irregular) 
%            and (hasAge some hasAge_veryHigh) and (hasBiRads some hasBiRads_high) SubClassOf ToLearn	0.965068

%  (hasDensity some low) and (hasMargin some microlobulated) 
%            and (hasShape some oval) and (hasBiRads some hasBiRads_medium) SubClassOf FALSEP_ToLearn	0.75

% hasBiRads_medium    triangular	     2.780   3.997       5.021654
% hasBiRads_high      right-shoulder  3.997   5.021654	
% hasAge_veryHigh     right-shoulder  62.793	71.882
% \end{verbatim}
% }
\end{example}

%\francacom{TABELLA: i calcoli non mi tornano se faccio F1 per penultima riga non mi viene 0,464 ma 0,469}
\begin{table}[H]        
%\begin{sidewaystable}

\caption{Results table. The measures are the macro average over the 5 folds \wrt~the test set.} \label{tab:allres}
\begin{center}
{\small
    \begin{tabular}{|c|l|c|c|c|c|}
    \hline
        \textbf{Dataset} & \textbf{Algorithm}  & \textbf{Precision} & \textbf{Recall} & \textbf{F1} & \textbf{\% Improvement } \\ \hline
        \multirow{2}{*}{\textbf{{\tt NTN}}} & Fuzzy DL-FOIL &  0.661 & 0.513 & 0.548 & \multirow{2}{*}{\textbf{80.47\%}}  \\ 
        \textbf{} & PN-OWL &  \textbf{1.000} & \textbf{0.980} & \textbf{0.989} &   \\ \hline \hline
        \multirow{2}{*}{\textbf{{\tt Lymphography}}} & Fuzzy DL-FOIL &  \textbf{0.861} & \textbf{0.851} & \textbf{0.855} &  
        \multirow{2}{*}{\textbf{-2.57\%}}  \\ 
        \textbf{} & PN-OWL &  0.836 & 0.841 & 0.833 &  \\ \hline\hline
         \multirow{2}{*}{\textbf{{\tt Mammographic}}} & Fuzzy DL-FOIL &  0.737 & 0.692 & 0.710 & 
         \multirow{2}{*}{\textbf{11.27\%}}  \\ 
        \textbf{} & PN-OWL &  \textbf{0.746} & \textbf{0.831} & \textbf{0.790} &    \\ \hline\hline
        \multirow{2}{*}{\textbf{{\tt Malware}}} & Fuzzy DL-FOIL &  0.623 & \textbf{0.830} & 0.704 & \multirow{2}{*}{\textbf{5.06\%}}   \\ 
        \textbf{} & PN-OWL & \textbf{0.701} & 0.818 & \textbf{0.740} & \\ \hline    \hline \hline
        \multirow{2}{*}{\textbf{{\tt Iris}}} & Fuzzy DL-FOIL &  0.886 & 0.910 & 0.890 & \multirow{2}{*}{\textbf{4.16\%}} \\ 
        \textbf{} & PN-OWL &  \textbf{0.949} & 0.910 & \textbf{0.927} &   \\ \hline\hline
        \multirow{2}{*}{\textbf{{\tt Wine}}} & Fuzzy DL-FOIL &  0.884 & \textbf{0.971} & 0.895 &  \multirow{2}{*}{\textbf{0.98\%}} \\ 
        \textbf{} & PN-OWL &  \textbf{0.933} & 0.904 & \textbf{0.914} &   \\ \hline\hline
        \multirow{2}{*}{\textbf{{\tt Wine Quality}}} & Fuzzy DL-FOIL &  0.227 & \textbf{0.917} & 0.363 & \multirow{2}{*}{\textbf{27.93\%}}   \\ 
        %\multirow{2}{*}{\textbf{{\tt Wine Quality}}} & Fuzzy DL-FOIL &  0.227 & \textbf{0.917} & 0.363 & \multirow{2}{*}{\textbf{29.20\%}}   \\ 
%
        \textbf{} & PN-OWL &  \textbf{0.365} & 0.659 & \textbf{0.464} &  \\ \hline\hline
%        \textbf{} & PN-OWL &  \textbf{0.365} & 0.659 & \textbf{0.469} &  \\ \hline\hline
        %
        \multirow{2}{*}{\textbf{{\tt YEAST}}} & Fuzzy DL-FOIL &  0.427 & 0.746 & 0.540 &  \multirow{2}{*}{\textbf{4.37\%}}  \\ 
        \textbf{} & PN-OWL & \textbf{0.432} & \textbf{0.815} & \textbf{0.564} & \\ \hline\hline
    \end{tabular}

} 
\end{center}
%\end{sidewaystable}
\end{table}

%----------------------------------
\vspace{1ex}
\nd {\bf Discussion.}
%\subsection{Discussion}
%---------------------------
%
\nd In Table~\ref{tab:allres},  the last column reports  the improvement of \fuzzyowlpnrule~relative to the measure $F1$ (see Eq.~\ref{f1eq}), over our baseline Fuzzy \foildl.
Overall, \fuzzyowlpnrule~performs  better than Fuzzy \foildl~(with the exception of {\tt Lymphography}) and in some cases the improvement is particularly high, such as for {\tt NTN, Mammographic} and {\tt Wine Quality}.

Essentially, for \fuzzyowlpnrule~we were able to find a better compromise between precision and recall than for \foildl. In particular, we were able to increase precision confirming our conjecture that indeed the N-stage is able to remove the false positives.
%(\eg~for NTN over all five runs, only one example in total has been mis-classified as false non-positive). 
%In general, the closer the $F1$ measure of \foildl~is to $1.0$ the less room there is for improvement.

Concerning {\tt Lymphography}, we were unable to replicate the results of  Fuzzy \foildl~in~\cite{Cardillo22}, for which we get now an F1 measure of 0.805 in place of 0.855. The difference lies in few miss-classified  examples. 
% We conjecture this may due to a slightly difference due to the 5-fold partitioning, even if in theory this should not have been the case as we have used the same random seeds. 
We also noted that in this case \fuzzyowlpnrule~achieves $F1=1.0$ during the training phase, which may suggest an over-fitting problem.

Last but not least, let us mention that  \fuzzyowlpnrule~(so does Fuzzy \foildl) does definitely not yet behave well on the  {\tt Wine Quality} and {\tt Yeast} datasets, which will be the subject of further investigation.

The overall lesson learnt with \fuzzyowlpnrule~is that indeed the N-stage may provide a non negligible contribution to improve effectiveness of the classification process, provided one may find the appropriate balance among precision and recall. Unfortunately, searching the parameter space of \fuzzyowlpnrule~for an optimum is quite time consuming and a brute-force approach may likely not be feasible (at least not with our computational resources at hand). In fact, we proceeded one run per time, and by analysing the results tried to figure out whether and how to change some of the parameters in Table~\ref{tab:param} to increase recall and/or precision. On the other-hand, optimising \foildl~is much easier as it has half of the parameters of \fuzzyowlpnrule.

%---------------------
\section{Related Work}
\label{sec:relatedWork}

\nd Concept inclusion axiom learning in DLs is essentially inspired by statistical relational learning, where classification rules are (possibly weighted) Horn clause theories (see \eg~\cite{deRaedt08,DeRaedt17}), and various methods have been proposed in the DL context so far (see \eg~\cite{Lisi19,dAmato20,Rettinger12}). The general idea consists in the exploration of the search space of potential concept descriptions that cover the available training examples using so-called refinement operators (see, \eg~\cite{Badea00,Chitsaz12,Lehmann09,Lehmann07,Lehmann07a,Lehmann10,Lisi03}).
The  goal is then to learn a concept description of the underlying DL language covering (possibly) all the provided positive examples and (possibly) not covering any of the provided negative examples. The fuzzy case (see~\cite{Lisi13,Lisi15,Straccia15,Cardillo22}) is a natural extension relying on fuzzy DLs~\cite{Bobillo15b,Straccia13} and fuzzy ILP (see \eg~\cite{SerrurierP07}) instead. 

As already mentioned, our two-stage algorithm is conceptually inspired by PN-rule~\cite{Agarwal00,Agarwal01,Joshi01,Joshi02} consisting of a P-stage in which positive rules (called P-rules) are learnt to cover as many as possible instances of a target class and an N-stage in which negative rules (called N-rules) are learnt to remove most of the non-positive examples covered by the P-stage. The two rule sets are then used to build up a decision method to classify an object being instance of the target class or not~\cite{Agarwal00,Agarwal01,Joshi01,Joshi02}. It is worth noting that what differentiates this method from all others is its second stage. The main differences of \fuzzyowlpnrule~\wrt~PN-rule are:
\ii{i} PN-rule operates with \emph{tabular data} only, \ie~the data consists of attribute value pairs $(A,v)$, while we are in the context of OWL ontologies.\footnote{Tabular data can easily be mapped into OWL ontologies as illustrated in~\cite{Cardillo22}.};  PN-rules are of the form $cond \rightarrow T$, where the condition $cond$ is of the form $(A \in [l,h])$ or $(A \not \in [l,h])$ for continuous attribute $A$.\footnote{If $A$ is categorical then obviously $cond$ is either of the form $A=v$ or $A\neq v$.}, while we have,  conjunction of conditions in the rule body and each condition may be fuzzy, besides being either a class name or a restriction on attributes (attributes may be also nested); and \ii{iii} PN-rule considers a completely different rule scoring and combination strategy than we use in \fuzzyowlpnrule. The latter can be represented in Fuzzy OWL 2~\cite{Bobillo10,Bobillo11c}, while for the former we conjecture it cannot: so, we left this option out as a fuzzy DL reasoner would not be able to reason with those types of rules.

Other closely related works are~\cite{Fanizzi08,Fanizzi11,Fanizzi18,Fanizzi19,Lisi13,Lisi15,Straccia15}. In fact, \cite{Fanizzi08,Fanizzi11,Fanizzi18,Rizzo20} can be seen as an adaption to the DL case of the the well-known  \foil-algorithm, while~\cite{Lisi13,Lisi15} that  stem essentially from~\cite{Lisi13a,Lisi14a,Lisi11a,Lisi11,Lisi13b,Lisi14}, propose \emph{fuzzy}  \foil-like algorithms instead, and are inspired by  fuzzy ILP variants such as~\cite{Drobics03,SerrurierP07,Shibata99}.\footnote{See, \eg~\cite{Cintra13}, for an overview on fuzzy rule learning methods.} Let us note that~\cite{Lisi13,Lisi11} consider the weaker hypothesis representation language \dllite~\cite{Artale09}, while  here we rely on an aggregation of fuzzy $\el(\D)$ inclusion axioms. Fuzzy $\el(\D)$ has also been considered in~\cite{Straccia15}, which however differs from~\cite{Lisi13,Lisi15} by the fact that a (fuzzy) probabilistic ensemble evaluation of the fuzzy concept description candidates has been considered.\footnote{Also, to the best of our knowledge, concrete datatypes were not addressed in the evaluation.} Let us recap that, to our opinion, fuzzy $\el(\D)$  concept expressions are appealing as they can straightforwardly be translated into natural language and, thus, contribute to the explainability aspect of the induced classifier.

Discrete boosting has been considered in~\cite{Fanizzi19} that also shows how  to derive a weak learner (called \textsc{wDLF}) from conventional learners using some sort of random downward  refinement operator covering at least a positive example and yielding a minimal score fixed with a threshold. Related to this work is~\cite{Cardillo22} that deals with fuzziness in the hypothesis language and a real-valued variant of AdaBoost and differentiates from the previous one by using a descent-like gradient algorithm to search for the best alternative. Notably, this  also deviates from `fuzzy' rule learning AdaBoost variants, such as~\cite{Jesus04,Otero06,Palacios11,Sanchez07,Zhu16} in which the weak learner is required to generate the whole rules' search space beforehand the selection of the best current alternative. 
Such an approach is essentially unfeasible in the OWL case due to the size of the search space.

In~\cite{Iglesias11} a method is described  that can learn fuzzy OWL DL concept equivalence axioms from FuzzyOWL 2 ontologies, by interfacing with the \emph{fuzzyDL} reasoner~\cite{Bobillo16}. The candidate concept expressions are provided by the underlying \textsc{DL-Learner}~\cite{Lehmann09a,Buehmann16,Buehmann18} system. However, it has been tested only on a toy ontology so far.
Moreover, let us mention~\cite{Konstantopoulos10} that is based on an ad-hoc translation of fuzzy \L{}ukasiewicz $\cal ALC$ DL constructs into fuzzy \emph{Logic Programming} (fuzzy LP) and uses a conventional  ILP method to learn rules. Unfortunately, the method is not sound as it has been shown that the mapping from fuzzy DLs to LP is incomplete~\cite{Motik07} and entailment in \L{}ukasiewicz $\cal ALC$ is undecidable \cite{Cerami13}. To be more precise, undecidability holds already for $\el$~under the infinitely valued \L{}ukasiewicz semantics~\cite{Borgwardt17a}.\footnote{We recall that $\el$ is a strict sub-logic of $\alc$.}

While it is not our aim to provide an extensive overview about learning \wrt~ontologies literature, we nevertheless recap here that there are also alternative methods to what we present here, but are related only to the extent that they deal with concept description induction in the context of DLs.
So, \eg, the series of works \cite{Fanizzi10,Fanizzi10a,Rizzo14,Rizzo14a,Rizzo15,Rizzo15b,Rizzo16a,Rizzo17,Rizzo18} are inspired on \emph{Decision Trees/Random Forests},
 \cite{Bloehdorn07,Fanizzi06,Fanizzi08a,Fanizzi12a} consider \emph{Kernel Methods} for inducing concept descriptions, while \cite{Minervini11,Minervini12,Minervini12a,Minervini14,Zhu15}  consider essentially a \emph{Naive Bayes} approach. Last but not least, \cite{Lehmann07b} is inspired on \emph{Genetic Programming} to induce concept expressions, while \cite{Nickles14} is based on the \emph{Reinforcement Learning} framework. Eventually, \cite{Rizzo21} proposes to use decision trees to learn so-called \emph{disjointness axioms}, \ie~expressions of the form $C\andc D \impc \bottomc$, declaring that class $C$ and $D$ are disjoint.

%%%%%%%%%%%%%%%%%%%%%%%%%%%%%%%%%%%%%%%%%%%%%%%%%%%%%%%%%%%%%%%%%%%%

\section{Conclusions \& Future Work}
\label{sec:conclusions}

\nd In this work, we  addressed the problem of automatically learning fuzzy concept inclusion axioms from OWL 2 ontologies to describe sufficient condition of being an individual classified as instance of target class $T$. That is, given a  target class $T$ of an OWL ontology, we have addressed the problem of inducing fuzzy concept inclusion axioms that describe sufficient conditions for being an individual instance  of $T$. 
Specifically, we have presented a two-stage algorithm, called \fuzzyowlpnrule~that is inspired on the PN-rule~\cite{Agarwal00,Agarwal01,Joshi01,Joshi02} and adapted to the context of OWL.
The main features of our algorithm are essentially the fact that
\ii{i} at the P-stage, it generates a set of fuzzy inclusion axioms, the P-rules, that cover as many as possible instances of the target class $T$ without compromising too much the amount on non-positives; 
\ii{ii} at the N-stage, it generates a set of fuzzy inclusion axioms, the N-rules, that cover as many as possible of non-positive instances of class $T$ of the P-stage; 
\ii{iii} the fuzzy inclusion axioms are then combined (aggregated) into a new fuzzy inclusion axiom describing sufficient conditions for being an individual classified as an instance of the target class $T$.
Additionally, all fuzzy inclusion axioms may possibly include fuzzy concepts and fuzzy concrete domains, where each axiom has a leveraging weight (specifically, called  confidence or precision), and all generated fuzzy concept inclusion axioms can directly be encoded as \emph{Fuzzy OWL 2} axioms.

We have also conducted an extensive evaluation, comparing it with fuzzy \foildl. Our evaluation shows that, \fuzzyowlpnrule~performs generally better than fuzzy \foildl~in terms of effectiveness, though finding an optimal parameter configuration is much more time consuming than for  \foildl~as \fuzzyowlpnrule~has double as many parameters than fuzzy \foildl.

Concerning future work, besides investigating about other learning methods, and future work listed here and there in the paper, we envisage various aspects worth to be investigated in more detail: 
\ii{i} it is still unclear how the construction of fuzzy sets may impact effectiveness. So far, we did not notice a clear winner between the uniform clustering and c-means clustering algorithms used to build fuzzy datatypes. This is somewhat surprising. We would like to investigate that in more detail by considering various alternatives as well~\cite{fssclustering20} and/or considering clustering methods based on the aggregation of data properties, \ie~multi-dimensional clustering versus uni-dimensional clustering;  
\ii{ii} moreover, we would like to cover more OWL datatypes than those considered here so far (numerical and boolean) such as strings, dates, etc., possibly in combination with some classical machine learning methods (see, \eg~\cite{ShalevShwartz14}); 
\ii{iii} we would like to investigate the computational aspect: so far, for some ontologies, a learning run may take even a week (\wrt~our available resources). Here, we would like to investigate both parallelization methods as well as to investigate about the impact, in terms of effectiveness, of efficient, logically sound, but not necessarily complete, reasoning algorithms; 
\ii{iv} in principle, our two-stage algorithm \fuzzyowlpnrule~is parametric \wrt~the learner to be used during both the P-stage and the N-stage (\cf~lines 5 and 19 of the \fuzzyowlpnrule~algorithm): here we would like to investigate how to plug in another alternative such as \fuzzyowladaboost~\cite{Cardillo22} and to verify its effectiveness; 
\ii{v} we would like to asses also the impact of other alternative scoring functions to information gain (\cf~Eq.~\ref{eq:gain}) within our setting, inclusive various alternative choices of t-norms and r-implications; and
$(vi)$ we are looking for combining our Fuzzy DL-Learning with sub-symbolic learning methods, such as \eg~Neural Networks, an activity that is already on-going. 

Moreover, we really would like to consider extending the hypothesis language $\eld$ with so-called \emph{threshold concepts}~\cite{Bobillo08a} of the form $C[\geq d]$ (resp.~$C[\leq d]$), where $d \in [0,1]$ and $C$ is either a class name or an existential restriction, with the intended meaning ``$C[\geq d]$ (resp.~$C[\leq d]$) is the fuzzy set of individuals that are instances of $C$ to degree greater (resp.~smaller) than or equal to $d$."
This would provide us a more fine grained hypothesis language in which a threshold may be defined for each conjunct of a rule rather than via a rule confidence threshold as it is now. A Fuzzy OWL 2 example of such a rule may be, by referring to the {\tt Wine Quality} ontology and target wine {\tt 1}
\begin{center}
%{\tiny
%{\scriptsize
{\footnotesize
\begin{verbatim}
(implies (and (some alcohol alcohol_VH)[<= 0.786] 
              (some sulphates sulphates_H)[>= 0.289]
              (some pH pH_L)[<= 0.106])
              1)
\end{verbatim}
}
\end{center}

\nd with intended meaning "if, for an individual (wine) $a$, the alcohol level of being very high is smaller than or equal to $0.786$, the sulphates level of being  high is greater than or equal to $0.289$ and the pH level of being low is smaller than or equal $0.106$ then classify $a$ to some extend (\eg~the minimum of the degrees of being $a$ an instance of a conjunct)  as instance of the target class {\tt 1}.

%------------------------------------------------------

\section*{Acknowledgment}
\nd This research was partially supported by TAILOR, a project funded by EU Horizon 2020 research and innovation programme under (GA No 952215). This work has also been partially supported by the H2020 DeepHealth Project (GA No. 825111). This paper is also supported by the FAIR (Future Artificial Intelligence Research) project funded by the NextGenerationEU program within the PNRR-PE-AI scheme (M4C2, investment 1.3, line on Artificial Intelligence). Eventually, this work has also been partially supported by the H2020 STARWARS Project (GA No. 101086252), type of action HORIZON TMA MSCA Staff Exchanges.

We wish to thank Centro Servizi CNR of the ICT-SAC Department of the
National Research Council for the precious computing services and
resources they made available. We wish to address a special thanks to
Ing. Giorgio Bartoccioni (ICT-SAC) for his technical support.

%------------------------------------------------------

{\footnotesize
%\bibliographystyle{plain}
%\bibliography{mybiblio}

}

% --------------------------------------
% APPENDIX
 \appendix

\section{Brief Description of the Datasets}\label{ontodesc}

%-------
%\subsection{OWL Ontologies Description}

\nd Find below a brief description about the OWL ontologies in Table~\ref{tab:onto} used in our experiments. 

\begin{description}
% \item[FamilyTree.] This is a simple family relationships ontology and associated instances. The description is of the family of Robert Stevens and the intention is to use the minimal of asserted relationships and the maximum of inference. To do this,  role chains, nominals and properties hierarchies have been used. The target is to identify sufficient conditions for being an uncle.

% \item[Hotel.] This ontology describes the meaningful entities of a city. Instances are hotels located in the town Pisa and ratings have been gathered from Trip Advisor.\footnote{\url{http://www.tripadvisor.com}} The target is to identify sufficient conditions for being a good hotel, which has been identified as a hotel having a rating above 4.

% \item[Moral.]  This ontology is about meaningful  entities involved in the description of guiltiness within a moral  theory of blame scenario. The target is to learn sufficient conditions to be guilty.

\item[SemanticBible (NTN).] \emph{New Testament Names} (NTN) is an ontology describing each named thing in the New Testament, about 600 names in all. Each named thing (an entity) is categorized according to its class, including God, Jesus, individual men and women, groups of people, and locations. These entities are related to each other by properties that interconnect the entities into a web of information.\footnote{\url{http://semanticbible.com/ntn/ntn-overview.html}} The target is to learn sufficient conditions to be a woman.

% \item[UBA.] This is a well-known  university ontology for benchmark tests describing meaningfull entities within a university (\eg~universities, departments and the activities that occur at them).\footnote{\url{http://swat.cse.lehigh.edu/projects/lubm/}} The target here is to determine sufficient conditions to be a good researcher.

% \item[WineOnto.] This is an ontology about Italian, French and German red and white wines involving the description of, among others, their chemical properties. The target here is to determine sufficient conditions to be a dry wine.

% \item[Pair50.] This ontology is about a poker game and the target is to determine whether a player has a pair at hand.

% \item[Straight.] This ontology is about a poker game, as the one for Pair50, but the target is now to determine whether one has a straight at hand.

\item[Lymphography.] This ontology is about lymphography patient data and the target is the prediction of a diagnosis class based on the lymphography patient data~\cite{Westphal19}.

\item[Mammographic.] This ontology is about mammography screening data and the target is the prediction of breast cancer severity based on the screening data~\cite{Westphal19}.

% \item[Pyrimidine.] This ontology is about pyrimidine data, the target is the prediction of the inhibition activity of pyrimidines and the DHFR enzyme~\cite{Westphal19}.

% \item[Suramin.] This ontology is about the description of chemical compounds and the target is to find a predictive description of suramin analogues for cancer treatment.

\item[Malware.] This ontology is the description of a PE Malware Ontology that offers a reusable semantic schema for Portable Executable (PE,Windows binary format) malware files~\cite{svec21,svec23}. The ontology is inspired by the structure of the data in the EMBER dataset,\footnote{https://github.com/elastic/ember} which is intended for static malware analysis~\cite{Anderson18}.

\end{description}

% \subsection{UCI ML Data Sets} \label{ucidesc}

 \nd The following datasets have been taken from the well-known \emph{UC Irvine Machine Learning Repository}~\cite{Dua:2019}. 

\begin{description}
\item[Iris.] The data set contains 3 classes of 50 instances each, where each class refers to a type of iris plant. The attributes are: 
sepal length in cm,
sepal width in cm,
petal length in cm and
petal width in cm. 
The target classes are: Iris Setosa, Iris Versicolour and Iris Virginica. 
%In this work we do not consider Iris setosa, as  

\item[Wine.] These data are the results of a chemical analysis of wines grown in the same region in Italy but derived from three different cultivars. The analysis determined the quantities of 13 constituents found in each of the three types of wines. The attributes are
alcohol, malic acid, ash, alcalinity of ash, magnesium, total phenols, flavonoids, nonflavonoid phenols, proanthocyanins, color intensity, hue, OD280/OD315 of diluted wines  and proline. The target classes are the three wines $1,2$ and $3$.

\item[Wine Quality.] The  data set is related to red and white variants of the Portuguese ``Vinho Verde" wine. 
The goal is to model wine quality based on physicochemical tests. 
Due to privacy and logistic issues, only physicochemical (inputs) and sensory (the output) variables are available (e.g. there is no data about grape types, wine brand, wine selling price, etc.). 
The attributes are: 
fixed acidity, volatile acidity, citric acid, residual sugar, chlorides, free sulfur dioxide, total sulfur dioxide, density, pH, sulphates, alcohol and quality (score between 0 and 10).
The target is to describe good red wines, which are defined as red wines having quality score greater than or equal to 7. The quality attribute has been removed from the ontology during training and tests.

\item[Yeast.] The data set is about the prediction of the cellular localization sites of proteins (10 target classes)
The set of attributes is:
Sequence Name (accession number for the SWISS-PROT database), 
mcg (McGeoch's method for signal sequence recognition);
gvh (von Heijne's method for signal sequence recognition);
alm (score of the ALOM membrane spanning region prediction program); 
mit (Score of discriminant analysis of the amino acid content of the N-terminal region, 20 residues long, of mitochondrial and non-mitochondrial proteins);
erl (presence of ``HDEL" substring, thought to act as a signal for retention in the endoplasmic reticulum lumen, binary attribute); 
pox (peroxisomal targeting signal in the C-terminus); 
vac (score of discriminant analysis of the amino acid content of vacuolar and extracellular proteins); and 
nuc (score of discriminant analysis of nuclear localization signals of nuclear and non-nuclear proteins).

\end{description}

%---------------------
%---------------------

\end{document}